\documentclass[10pt,twocolumn,letterpaper]{article}
\usepackage[accsupp]{axessibility}  % Improves PDF readability for those with disabilities.

\usepackage{cvpr}              % To produce the CAMERA-READY version
\usepackage{caption}
\usepackage{subcaption}
\usepackage{graphicx}
\usepackage{amsmath}
\usepackage{amssymb}
\usepackage{booktabs}
\usepackage{enumitem}

\usepackage[table]{xcolor}
\usepackage[pagebackref, breaklinks=true, colorlinks, citecolor=citecolor, linkcolor=linkcolor, bookmarks=false]{hyperref}
\definecolor{citecolor}{HTML}{0071BC}
\definecolor{linkcolor}{HTML}{ED1C24}

\usepackage[capitalize]{cleveref}
\crefname{section}{Sec.}{Secs.}
\Crefname{section}{Section}{Sections}
\crefname{table}{Tab.}{Tabs.}
\Crefname{table}{Table}{Tables}

\definecolor{poscolor}{RGB}{212,17,89}
\definecolor{color2}{RGB}{250,130,130}
\definecolor{color3}{RGB}{130,130,130}
\definecolor{color4}{RGB}{14,83,246}
\newcommand{\aaa}[1]{\textcolor{poscolor}{\footnotesize \textbf{(#1)}}}
\newcommand{\bbb}[1]{\textcolor{color2}{\footnotesize (#1)}}
\newcommand{\ccc}[1]{\textcolor{color3}{\footnotesize (#1)}}
\newcommand{\zzz}[1]{\textcolor{color4}{\footnotesize \textbf{(#1)}}}

\newcommand{\tabincell}[2]{\begin{tabular}{@{}#1@{}}#2\end{tabular}}
\usepackage{algorithm}
\usepackage{bm}
\usepackage{algorithmic}
\usepackage{listings}
\usepackage{makecell}  % \Xhline
\usepackage{diagbox} % diagonal table cell
\usepackage{mathtools} % \coloneqq :=
\usepackage{graphicx} % rotated text
\usepackage{multirow}

\def\ours{DiffusionDet\xspace}  % Submit

\definecolor{detcolor}{gray}{.9}
\newcommand{\diffcell}[1]{\cellcolor{detcolor}{#1}}

\definecolor{bestcolor}{gray}{.9}
\newcommand{\bestcell}[1]{\cellcolor{bestcolor}{#1}}

\newcommand{\tablestyle}[2]{\setlength{\tabcolsep}{#1}\renewcommand{\arraystretch}{#2}\centering\footnotesize}
\newlength\savewidth\newcommand\shline{\noalign{\global\savewidth\arrayrulewidth
  \global\arrayrulewidth 1pt}\hline\noalign{\global\arrayrulewidth\savewidth}}
  
\newcolumntype{x}[1]{>{\centering\arraybackslash}p{#1pt}}
\newcolumntype{y}[1]{>{\raggedright\arraybackslash}p{#1pt}}
\newcolumntype{z}[1]{>{\raggedleft\arraybackslash}p{#1pt}}
\renewcommand{\paragraph}[1]{\vspace{1.25mm}\noindent\textbf{#1}}
\definecolor{deemph}{gray}{0.6}
\newcommand{\gc}[1]{\textcolor{deemph}{#1}}
\usepackage{etoolbox}
\makeatletter
\AfterEndEnvironment{algorithm}{\let\@algcomment\relax}
\AtEndEnvironment{algorithm}{\kern2pt\hrule\relax\vskip3pt\@algcomment}
\let\@algcomment\relax
\newcommand\algcomment[1]{\def\@algcomment{\footnotesize#1}}
\renewcommand\fs@ruled{\def\@fs@cfont{\bfseries}\let\@fs@capt\floatc@ruled
  \def\@fs@pre{\hrule height.8pt depth0pt \kern2pt}%
  \def\@fs@post{}%
  \def\@fs@mid{\kern2pt\hrule\kern2pt}%
  \let\@fs@iftopcapt\iftrue}
\makeatother
\def\cvprPaperID{5600} % *** Enter the CVPR Paper ID here
\def\confName{ICCV}
\def\confYear{2023}

\begin{document}

%%%%%%%%% TITLE - PLEASE UPDATE
\title{\ours: Diffusion Model for Object Detection}

\author{Shoufa Chen$^1$ \quad Peize Sun$^1$ \quad Yibing Song$^{2,3}$ \quad Ping Luo$^{1,4}$\\
$^1$The University of Hong Kong \quad $^2$Tencent AI Lab \\ $^3$AI$^3$ Institute, Fudan University  \quad $^4$Shanghai AI Laboratory \\
{\tt\small \{sfchen, pzsun, pluo\}@cs.hku.hk \quad yibingsong.cv@gmail.com}
}
\maketitle

%%%%%%%%% ABSTRACT
\begin{abstract}
We propose DiffusionDet, a new framework that formulates object detection as a denoising diffusion process from noisy boxes to object boxes. During the training stage, object boxes diffuse from ground-truth boxes to random distribution, and the model learns to reverse this noising process. In inference, the model refines a set of randomly generated boxes to the output results in a progressive way. 
Our work possesses an appealing property of flexibility, which enables the dynamic number of boxes and iterative evaluation. The extensive experiments on the standard benchmarks show that DiffusionDet achieves favorable performance compared to previous well-established detectors. For example, DiffusionDet achieves 5.3 AP and 4.8 AP gains when evaluated with more boxes and iteration steps, under a zero-shot transfer setting from COCO to CrowdHuman. Our code is available at \url{https://github.com/ShoufaChen/DiffusionDet}.
\end{abstract}

\section{Introduction}\label{sec:intro}
Object detection aims to predict a set of bounding boxes and associated category labels for targeted objects in one image. As a fundamental visual recognition task, it has become the cornerstone of many related recognition scenarios, such as instance segmentation~\cite{he2017mask, li2017fully}, pose estimation~\cite{fang2017rmpe, openpose}, action recognition~\cite{sigurdsson2016hollywood,gu2018ava}, object tracking~\cite{kalal2011tracking, milan2016mot16}, and visual relationship detection~\cite{lu2016visual, johnson2015image}.

\begin{figure}[t]
\includegraphics[width=.98\linewidth]{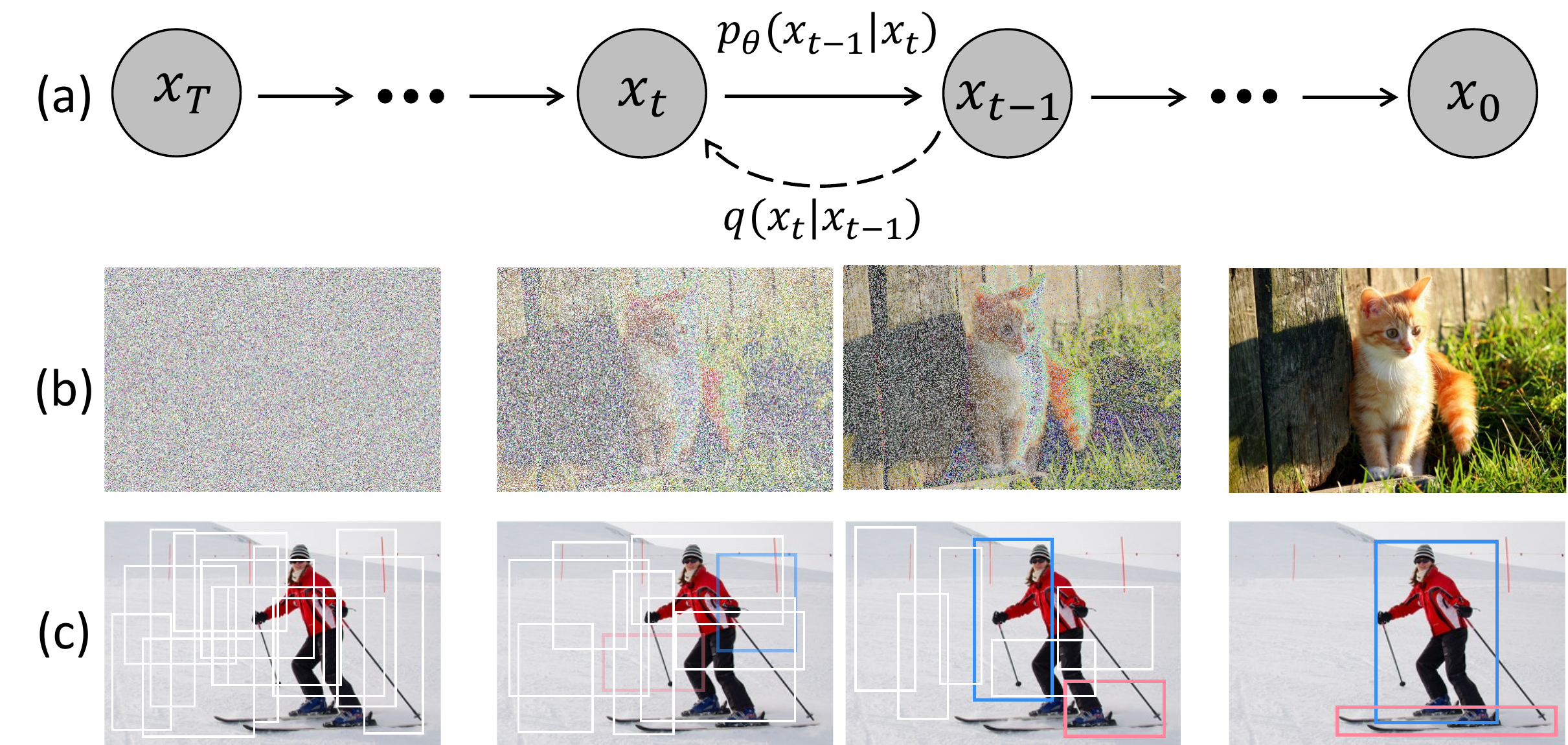}
\vspace{-5pt}
\caption{\textbf{Diffusion model for object detection}. (a) A diffusion model where $q$ is the diffusion process and $p_\theta$ is the reverse process. (b) Diffusion model for image generation task. (c) We propose to formulate object detection as a denoising diffusion process from noisy boxes to object boxes.
}
\label{fig:teaser}
\end{figure}

Modern object detection approaches have been evolving with the development of object candidates, \ie, from empirical object priors~\cite{girshick2015fast, ren2015faster, redmon2016you, liu2016ssd} to learnable object queries~\cite{carion2020end, zhu2021deformable, sun2021sparse}).
Specifically, the majority of detectors solve detection tasks by defining surrogate regression and classification on empirically designed object candidates, such as sliding windows~\cite{sermanet2013overfeat, girshick2014rich}, region proposals~\cite{girshick2015fast, ren2015faster}, anchor boxes~\cite{redmon2016you, lin2017focal} and reference points~\cite{zhou2019objects, duan2019centernet, yang2019reppoints}. Recently, DETR~\cite{carion2020end} proposes learnable object queries to eliminate the hand-designed components and set up an end-to-end detection pipeline, attracting great attention on query-based detection paradigm~\cite{zhu2021deformable, sun2021sparse, Gao_2021_ICCV, li2022dn}. 

While these works achieve a simple and effective design, they still have a dependency on a fixed set of learnable queries.
A natural question is: \emph{is there a simpler approach that does not even need the surrogate of learnable queries}? 

We answer this question by designing a novel framework that directly detects objects from a set of random boxes. Starting from purely random boxes, which do not contain learnable parameters that need to be optimized in the training stage, we expect to gradually refine the positions and sizes of these boxes until they perfectly cover the targeted objects. 
This \emph{noise-to-box} approach requires neither heuristic object priors nor learnable queries, further simplifying the object candidates and pushing the development of the detection pipeline forward.

Our motivation is illustrated in \Cref{fig:teaser}. We think of the philosophy of noise-to-box paradigm is analogous to \emph{noise-to-image} process in the denoising diffusion models~\cite{ho2020denoising, song2021scorebased, dhariwal2021diffusion}, which are a class of likelihood-based models to generate the image by gradually removing noise from an image via the learned denoising model. Diffusion models have achieved great success in many generation tasks~\cite{avrahami2022blended, Ramesh2022HierarchicalTI, austin2021structured, hoogeboom2022equivariant, trippe2022diffusion} and start to be explored in perception tasks like image segmentation~\cite{wolleb2021diffusion, baranchuk2022labelefficient, graikos2022diffusion, kim2022diffusion, brempong2022denoising, amit2021segdiff, chen2022generalist}. However, to the best of our knowledge, there is no prior art that successfully adopts it to object detection.

In this work, we propose \ours, which tackles the object detection task with a diffusion model by casting detection as a generative task over the space of the positions~(center coordinates) and sizes~(widths and heights) of bounding boxes in the image. At the training stage, Gaussian noise controlled by a variance schedule~\cite{ho2020denoising} is added to ground truth boxes to obtain \emph{noisy} boxes.
Then these noisy boxes are used to crop~\cite{ren2015faster,he2017mask} features of Region of Interest~(RoI) from the output feature map of the backbone encoder, \eg, ResNet~\cite{he2016deep}, Swin Transformer~\cite{liu2021swin}. Finally, these RoI features are sent to the detection decoder, which is trained to predict the ground-truth boxes without noise. With this training objective, \ours is able to predict the ground truth boxes from random boxes.  
At the inference stage, \ours generates bounding boxes by reversing the learned diffusion process, which adjusts a noisy prior distribution to the learned distribution over bounding boxes.

\begin{table}[t]
    \small
    \renewcommand{\arraystretch}{.96}
    \begin{subtable}[ht]{0.98\linewidth}
    \tabcolsep=0.14cm
    \centering
    \begin{tabular}{@{}lcccc@{}}
    \toprule
    \emph{\small \# Boxes} & 300 & 500 & 1000 & 2000  \\  \midrule
     DETR~\cite{carion2020end}   & 61.3 & 61.3~\ccc{+0.0} & 61.3~\ccc{+0.0} & 61.3~\ccc{+0.0} \\
     Sparse R-CNN~\cite{sun2021sparse} & 66.6 & 66.5~\zzz{-0.1} & 66.5~\zzz{-0.1} & 66.5~\zzz{-0.1} \\ 
    \ours & 66.6 & 69.0~\aaa{+2.4} & 71.0~\aaa{+4.4} & 71.9~\aaa{+5.3} \\
    \bottomrule
    \end{tabular}
    \vspace{-4pt}
    \caption{Dynamic number of evaluation \textbf{\emph{boxes}}.}
    \end{subtable}
    \begin{subtable}[ht]{0.98\linewidth}
    \vspace{4pt}
    \tabcolsep=0.14cm
    \centering
    \begin{tabular}{@{}lcccc@{}}
    \toprule
    \emph{\small \# Steps} & 1 & 2 & 3 & 4 \\  \midrule
    DETR~\cite{carion2020end}   & 61.3 & 62.5~\bbb{+1.2} & 62.7~\bbb{+1.4} & 62.7~\bbb{+1.4} \\
    Sparse R-CNN~\cite{sun2021sparse} & 66.6 & 60.6~\zzz{-6.0} & 55.5~\zzz{-11.1} & 52.6~\zzz{-14.0} \\
    \ours & 66.6 & 69.7~\aaa{+3.1} & 70.8~\aaa{+4.2} & 71.4~\aaa{+4.8} \\
    \bottomrule
    \end{tabular}
    \vspace{-4pt}
    \caption{Dynamic number of evaluation \textbf{\emph{steps}}.}
    \end{subtable}
\caption{\textbf{Zero-shot transfer from COCO to CrowdHuman visible box detection.} All models are trained with 300 boxes and tested with different number of boxes and steps.}\label{tab:zero-shot-crowdhuman}
\end{table}

As a probabilistic model, \ours has an attractive superiority of flexibility, \ie, we can train the network once and use the same network parameters under diverse settings in the inference stage, mainly including: (1)~\emph{Dynamic number of boxes.} Leveraging random boxes as object candidates, we decouple the training and evaluation stage of \ours, \ie, we can train \ours with $N_{train}$ random boxes while evaluating it with $N_{eval}$ random boxes, where the $N_{eval}$ is arbitrary and does not need to be equal to $N_{train}$. (2)~\emph{Iterative evaluation.} Benefited by the iterative denoising property of diffusion models, \ours can reuse the whole detection head in an iterative way, further improving its performance.

The flexibility of \ours makes it a great advantage in detecting objects across different scenarios, \eg, sparse or crowded, without additional fine-tuning. Specifically, \Cref{tab:zero-shot-crowdhuman} shows that when directly evaluating COCO-pretraiend models on CrowdHuman~\cite{shao2018crowdhuman} dataset, which covers more crowed scenes, \ours achieves significant gains by adjusting the number of evaluation boxes and iteration steps. In contrast, previous methods only obtain marginal gains or even degraded performance. More detailed discussions are left in \Cref{sec:exp}.

Besides, we evaluate \ours on COCO~\cite{lin2014microsoft} dataset.
With ResNet-50~\cite{he2016deep} backbone, \ours achieves 45.8 AP using a single sampling step and 300 random boxes, which significantly outperforms Faster R-CNN~\cite{ren2015faster}~(40.2 AP), DETR~\cite{carion2020end}~(42.0 AP) and on par with Sparse R-CNN~\cite{sun2021sparse}~(45.0 AP). Besides, we can further improve \ours up to 46.8 AP by increasing the number of sampling steps and random boxes.

\noindent
Our \textbf{contributions} are summarized as follows:
\begin{itemize}[noitemsep,nolistsep,leftmargin=*]
    \item We formulate object detection as a generative denoising process, which is the first study to apply the diffusion model to object detection to the best of our knowledge.    
    \item Our noise-to-box detection paradigm has several appealing properties, such as decoupling training and evaluation stage for dynamic boxes and iterative evaluation.
    \item We conduct extensive experiments on COCO, CrowdHuman, and LVIS benchmarks. \ours achieves favorable performance against previous well-established detectors, especially zero-shot transferring across different scenarios.
\end{itemize}

\section{Related Work}\label{sec:related_work}

\vspace{-3pt}
\paragraph{Object detection.}
Most modern object detection approaches perform box regression and category classification on empirical object priors, such as proposals~\cite{ren2015faster, girshick2015fast}, anchors~\cite{redmon2016you, redmon2017yolo9000, lin2017focal}, points~\cite{zhou2019objects, tian2019fcos, wang2017point}. Recently, Carion~\etal proposed DETR~\cite{carion2020end} to detect objects using a fixed set of learnable queries. Since then, the query-based detection paradigm has attracted great attention and inspired a series of following works~\cite{zhu2021deformable, sun2021sparse, sun2021makes, meng2021conditional, liu2022dabdetr, li2022dn, gao2022adamixer, zhang2022expected, zhang2022dino, jia2022detrs, chen2022group, nguyen2022boxer}. In this work, we push forward the development of the object detection pipeline further with \ours.

\paragraph{Diffusion model.}
As a class of deep generative models, diffusion models~\cite{ho2020denoising, song2019generative, song2021scorebased} start from the sample in random distribution and recover the data sample via a gradual denoising process. 
Diffusion models have recently demonstrated remarkable results in fields including 
computer vision~\cite{avrahami2022blended, Ramesh2022HierarchicalTI, saharia2022photorealistic, pmlr-v162-nichol22a, gu2022vector, Fan2022FridoFP, Ruiz2022DreamBoothFT, singer2022make, harvey2022flexible, zhang2022motiondiffuse, ho2022video, yang2022diffusion},
nature language processing~\cite{austin2021structured, li2022diffusion, gong2022diffuseq},
audio processing~\cite{Popov2021GradTTSAD, yang2022diffsound, wu2021itotts, levkovitch2022zero, tae2021editts, huang2022prodiff, kim2022guided}, 
graph-related topics~\cite{jang2023diffusion},
interdisciplinary applications\cite{jing2022torsional, hoogeboom2022equivariant, anand2022protein, xu2021geodiff, trippe2022diffusion, wu2022diffusion, arne2022structure}, etc. More applications of diffusion models can be found in recent surveys~\cite{yang2022diffusion, cao2022survey}.

\paragraph{Diffusion model for perception tasks.}
While Diffusion models have achieved great success in image generation~\cite{ho2020denoising, song2021scorebased, dhariwal2021diffusion}, their potential for discriminative tasks has yet to be fully explored. Some pioneer works tried to adopt the diffusion model for image segmentation tasks~\cite{wolleb2021diffusion, baranchuk2022labelefficient, graikos2022diffusion, kim2022diffusion, brempong2022denoising, amit2021segdiff, chen2022generalist}, for example, Chen~\etal~\cite{chen2022generalist} adopted Bit Diffusion model~\cite{chen2022analog} for panoptic segmentation~\cite{kirillov2019panoptic} of images and videos. However, despite significant interest in this idea,  
there are no previous solutions that successfully adapt generative diffusion models for object detection, the progress of which remarkably lags behind that of segmentation. We argue that this may be because segmentation tasks are processed in an image-to-image style, which is more conceptually similar to the image generation tasks, while object detection is a set prediction problem~\cite{carion2020end} which requires assigning object candidates~\cite{ren2015faster,lin2017feature,carion2020end} to ground truth objects.
To the best of our knowledge, this is the first work that adopts a diffusion model for object detection.
\section{Approach}

\subsection{Preliminaries}

\paragraph{Object detection.}
The learning objective of object detection is input-target pairs $(\bm{x}, \bm{b}, \bm{c})$, where $\bm{x}$ is the input image, $\bm{b}$ and $\bm{c}$ are a set of bounding boxes and category labels for objects in the image $\bm{x}$, respectively. More specifically, we formulate the $i$-th box in the set as $\bm{b}^i = (c^i_x, c^i_y, w^i, h^i)$, where $(c^i_x, c^i_y)$ is the center coordinates of the bounding box, $(w^i, h^i)$ are width and height of that bounding box, respectively. 

\begin{figure}[t]
    \centering
    \begin{subfigure}{0.98\linewidth}
        \centering
        \includegraphics[width=0.72\linewidth]{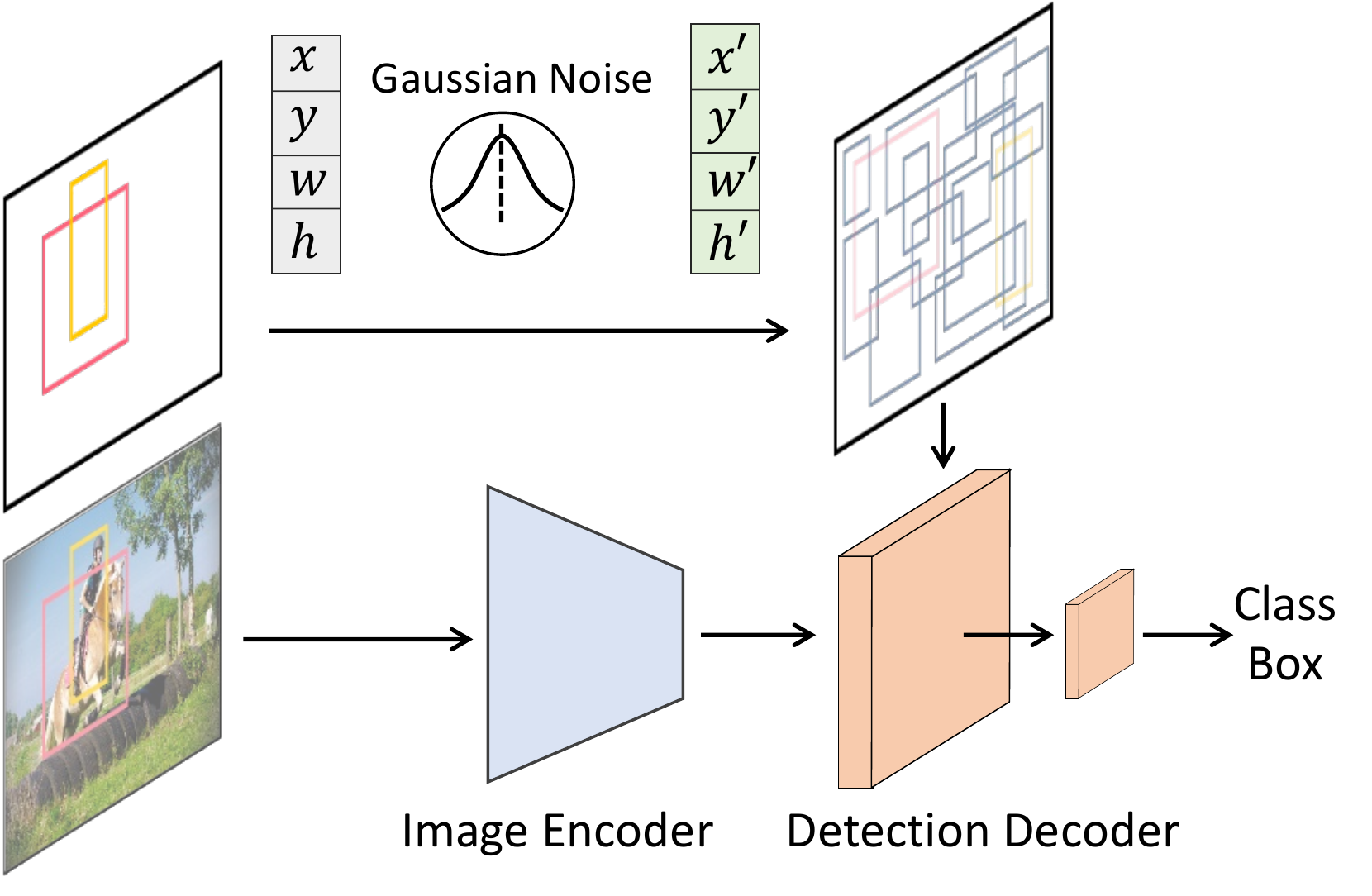}
        \caption{Overall pipeline.}\label{fig:framework}
    \end{subfigure}
    \begin{subfigure}{0.98\linewidth}
        \centering
        \includegraphics[width=0.96\linewidth]{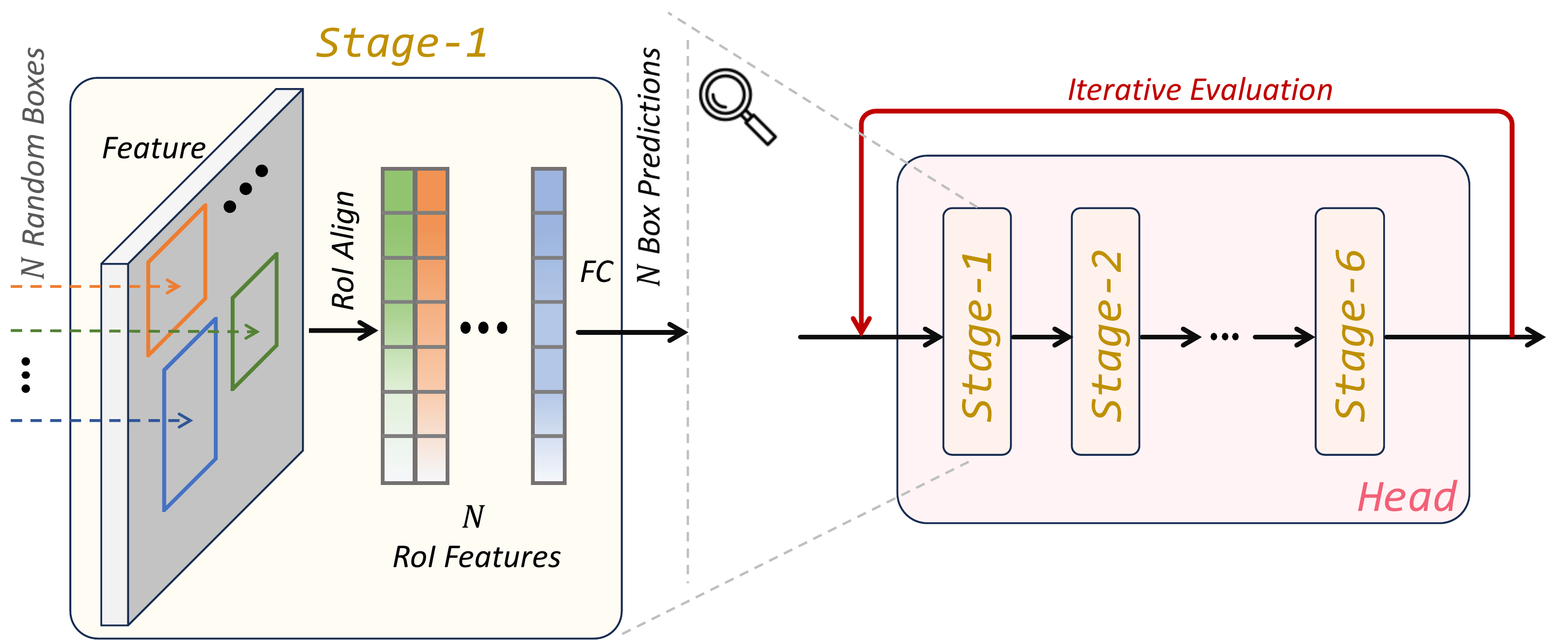}
        \caption{Details of the detection decoder/head.}
    \label{fig:head_details}
    \end{subfigure}
    \caption{\textbf{\ours framework.}
    \textbf{(a)} The image encoder extracts feature representation from an input image. The detection decoder takes noisy boxes as input and predicts category classification and box coordinates.
    \textbf{(b)} The detection decoder has 6 stages in one detection \texttt{head}, following DETR and Sparse R-CNN. Besides, DiffusionDet can reuse this detection head~(with 6 stages) multiple times, which is called \emph{``iterative evaluation''}.
 }
\end{figure}

\paragraph{Diffusion model.}
Diffusion models~\cite{sohl2015deep, ho2020denoising, song2019generative, song2021denoising} are a classes of likelihood-based models inspired by nonequilibrium thermodynamics~\cite{song2019generative, song2020improved}. These models define a Markovian chain of diffusion forward process by gradually adding noise to sample data. The forward noise process is defined as

\begin{equation}
\label{eq:noise_process}
    q(\bm{z}_t | \bm{z}_0) = \mathcal{N}(\bm{z}_t | \sqrt{\bar{\alpha}_t} \bm{z}_0, (1 - \bar{\alpha}_t) \bm{I}),
\end{equation}
which transforms data sample $\bm{z}_0$ to a latent noisy sample $\bm{z}_t$ for $t\in\{0, 1, ...,T\}$ by adding noise to $\bm{z}_0$.
$\bar{\alpha}_t \coloneqq \prod_{s=0}^{t} \alpha_s = \prod_{s=0}^{t} (1 - \beta_s)$ and $\beta_s$ represents the noise variance schedule~\cite{ ho2020denoising}.
During training, a neural network $f_\theta(\bm{z}_t, t)$ is trained to predict $\bm{z}_0$ from $\bm{z}_t$ by minimizing the training objective with $\ell_2$ loss~\cite{ho2020denoising}:
\begin{equation}
    \mathcal{L}_\text{train} =  \frac{1}{2}|| f_\theta(\bm{z}_t, t) - \bm{z}_0 ||^2.
\end{equation}
At inference stage, data sample $\bm{z}_0$ is reconstructed from noise $\bm{z}_T$ with the model $f_\theta$ and an updating rule~\cite{ho2020denoising, song2021denoising} in an iterative way, \ie,  $\bm{z}_T \rightarrow \bm{z}_{T-\Delta} \rightarrow ... \rightarrow \bm{z}_0$. More detailed formulation of diffusion models can be found in Appendix A.

In this work, we aim to solve the object detection task via the diffusion model. In our setting, data samples are a set of bounding boxes $\bm{z}_0 = \bm{b}$, where $\bm{b}\in \mathbb{R}^{N \times 4}$ is a set of $N$ boxes. 
A neural network $f_\theta(\bm{z}_t, t, \bm{x})$ is trained to predict $\bm{z}_0$ from noisy boxes $\bm{z}_t$, conditioned on the corresponding image $\bm{x}$. The corresponding category label $\bm{c}$ is produced accordingly. 

\subsection{Architecture} 
Since the diffusion model generates data samples iteratively, it needs to run model $f_\theta$ multiple times at the inference stage. However, it would be computationally intractable to directly apply $f_\theta$ on the raw image at every iterative step.
Therefore, we propose to separate the whole model into two parts, \textit{image encoder} and \textit{detection decoder}, where the former runs only once to extract a deep feature representation from the raw input image $\bm{x}$, and the latter takes this deep feature as condition, instead of the raw image, to progressively refine the box predictions from noisy boxes $\bm{z}_t$.

\paragraph{Image encoder.} Image encoder takes as input the raw image 
and extracts its high-level features for the following detection decoder. We implement \ours with both Convolutional Neural Networks such as ResNet~\cite{he2016deep} and Transformer-based models like Swin~\cite{liu2021swin}. Feature Pyramid Network~\cite{lin2017feature} is used to generate multi-scale feature maps for both ResNet and Swin backbones following~\cite{lin2017feature, sun2021sparse, liu2021swin}.

\paragraph{Detection decoder.} 
Borrowed from Sparse R-CNN~\cite{sun2021sparse}, the detection decoder takes as input a set of proposal boxes to crop RoI-feature~\cite{ren2015faster,he2017mask} from feature map generated by image encoder, and sends these RoI-features to detection head to obtain box regression and classification results.
For DiffusionDet, these proposal boxes are disturbed from ground truth boxes at training stage and directly sampled from Gaussian distribution at evaluation stage.
Following~\cite{carion2020end, zhu2021deformable, sun2021sparse}, our detection decoder is composed of 6 cascading stages~(\Cref{fig:head_details}). The differences between our decoder and the one in Sparse R-CNN are that (1) \ours begins from random boxes while Sparse R-CNN uses a fixed set of learned boxes in inference; (2) Sparse R-CNN takes as input pairs of the proposal boxes and its corresponding proposal feature, while \ours needs the proposal boxes only; (3) \ours can re-use the detector head in an iterative way for evaluation and the parameters are shared across different steps, each of which is specified to the diffusion process by timestep embedding~\cite{ho2020denoising}, which is called \emph{iterative evaluation}, while Sparse R-CNN uses the detection decoder only once in the forward pass.

\begin{algorithm}[t]
\small
\caption{\small \ours Training 
}
\label{alg:train}
\algcomment{\fontsize{7.2pt}{0em}\selectfont \texttt{alpha\_cumprod(t)}: cumulative product of $\alpha_i$, \ie, $\prod_{i=1}^t \alpha_i$
}
\definecolor{codeblue}{rgb}{0.25,0.5,0.5}
\definecolor{codegreen}{rgb}{0,0.6,0}
\definecolor{codekw}{RGB}{207,33,46}
\lstset{
  backgroundcolor=\color{white},
  basicstyle=\fontsize{7.5pt}{7.5pt}\ttfamily\selectfont,
  columns=fullflexible,
  breaklines=true,
  captionpos=b,
  commentstyle=\fontsize{7.5pt}{7.5pt}\color{codegreen},
  keywordstyle=\fontsize{7.5pt}{7.5pt}\color{codekw},
  escapechar={|}, 
}
\begin{lstlisting}[language=python]
def train_loss(images, gt_boxes):
  """
  images: [B, H, W, 3]
  gt_boxes: [B, *, 4]
  # B: batch
  # N: number of proposal boxes
  """
  
  # Encode image features
  feats = image_encoder(images)
  
  # Pad gt_boxes to N
  pb = pad_boxes(gt_boxes) # padded boxes: [B, N, 4]

  # Signal scaling
  pb = (pb * 2 - 1) * scale  

  # Corrupt gt_boxes
  t = randint(0, T)|~~~~~~~~~~~|# time step
  eps = normal(mean=0, std=1)  # noise: [B, N, 4]
  pb_crpt = sqrt(|~~~~|alpha_cumprod(t)) * pb + 
              |~|sqrt(1 - alpha_cumprod(t)) * eps

  # Predict
  pb_pred = detection_decoder(pb_crpt, feats, t)

  # Set prediction loss
  loss = set_prediction_loss(pb_pred, gt_boxes)
  
  return loss
\end{lstlisting}
\end{algorithm}

\subsection{Training}\label{sec:method_train}

During training, we first construct the diffusion process from ground-truth boxes to noisy boxes and then train the model to reverse this process.  \Cref{alg:train} provides the pseudo-code of \ours training procedure.

\paragraph{Ground truth boxes padding.}
For modern object detection benchmarks~\cite{everingham2010pascal, lin2014microsoft, shao2018crowdhuman, gupta2019lvis}, the number of instances of interest typically varies across images.  Therefore, we first \emph{pad} some extra boxes to original ground truth boxes such that all boxes are summed up to a fixed number $N_{train}$. 
We explore several padding strategies, for example, repeating existing ground truth boxes, concatenating random boxes or image-size boxes. 
Comparisons of these strategies are in \Cref{sec:exp_ablation}, and concatenating random boxes works best.

\paragraph{Box corruption.}
We add Gaussian noises to the padded ground truth boxes. The noise scale is controlled by $\alpha_t$~(in \cref{eq:noise_process}), which adopts the monotonically decreasing cosine schedule for $\alpha_t$ in different time step $t$, as proposed in~\cite{nichol2021improved}. Notably, the ground truth box coordinates need to be scaled as well since the signal-to-noise ratio has a significant effect on the performance of diffusion model~\cite{chen2022generalist}. We observe that object detection favors a relatively higher signal scaling value than image generation task~\cite{ho2020denoising, dhariwal2021diffusion, chen2022analog}. More discussions are in \Cref{sec:exp_ablation}.

\paragraph{Training losses.}
The detection detector takes as input $N_{train}$ corrupted boxes and predicts $N_{train}$ predictions of category classification and box coordinates. We apply set prediction loss~\cite{carion2020end, sun2021sparse, zhu2021deformable} on the set of $N_{train}$ predictions. We assign multiple predictions to each ground truth by selecting the top $k$ predictions with the least cost by an optimal transport assignment method~\cite{ge2021ota, ge2021yolox, wu2022defense, du20211st}. 

\subsection{Inference}\label{sec:inference}

The inference procedure of \ours is a denoising sampling process from noise to object boxes. Starting from boxes sampled in Gaussian distribution, the model progressively refines its predictions, as shown in \Cref{alg:sample}. 

\begin{algorithm}[t]
\small
\caption{\small \ours Sampling 
}
\label{alg:sample}
\algcomment{\fontsize{7.2pt}{0em}\selectfont \texttt{linespace}: generate evenly spaced values
}
\definecolor{codeblue}{rgb}{0.25,0.5,0.5}
\definecolor{codegreen}{rgb}{0,0.6,0}
\definecolor{codekw}{rgb}{0.85, 0.18, 0.50}
\lstset{
  backgroundcolor=\color{white},
  basicstyle=\fontsize{7.5pt}{7.5pt}\ttfamily\selectfont,
  columns=fullflexible,
  breaklines=true,
  captionpos=b,
  commentstyle=\fontsize{7.5pt}{7.5pt}\color{codegreen},
  keywordstyle=\fontsize{7.5pt}{7.5pt}\color{codekw},
  escapechar={|}, 
}
\begin{lstlisting}[language=python]
def infer(images, steps, T):
  """
  images: [B, H, W, 3]
  # steps: number of sample steps
  # T: number of time steps
  """
  
  # Encode image features
  feats = image_encoder(images)
  
  # noisy boxes: [B, N, 4]
  pb_t = normal(mean=0, std=1)

  # uniform sample step size
  times = reversed(linespace(-1, T, steps))
  
  # [(T-1, T-2), (T-2, T-3), ..., (1, 0), (0, -1)]
  time_pairs = list(zip(times[:-1], times[1:])

  for t_now, t_next in zip(time_pairs):
    # Predict pb_0 from pb_t
    pb_pred = detection_decoder(pb_t, feats, t_now)
    
    # Estimate pb_t at t_next
    pb_t = ddim_step(pb_t, pb_pred, t_now, t_next)

    # Box renewal
    pb_t = box_renewal(pb_t)
    
  return pb_pred
\end{lstlisting}

\end{algorithm}

\paragraph{Sampling step.}
In each sampling step, the random boxes or the estimated boxes from the last sampling step are sent into the detection decoder to predict the category classification and box coordinates. After obtaining the boxes of the current step, DDIM~\cite{song2021denoising} is adopted to estimate the boxes for the next step. We note that sending the predicted boxes without DDIM to the next step is also an optional progressive refinement strategy. However, it brings significant deterioration, as discussed in \Cref{sec:exp_ablation}.

\paragraph{Box renewal.}
After each sampling step, the predicted boxes can be coarsely categorized into two types, \emph{desired} and \emph{undesired} predictions. The desired predictions contain boxes that are properly located at corresponding objects, while the undesired ones are distributed arbitrarily. Directly sending these undesired boxes to the next sampling iteration would not bring a benefit
since their distribution is not constructed by box corruption in training.
To make inference better align with training, we propose the strategy of \emph{box renewal} to revive these undesired boxes by replacing them with random boxes.
Specifically, we first filter out undesired boxes with scores lower than a particular threshold. Then, we concatenate the remaining boxes with new random boxes sampled from a Gaussian distribution.

\paragraph{Flexible usage.}
Thanks to the random boxes design, we can evaluate \ours with an arbitrary number of random boxes and the number of iteration times, which do not need to be equal to the training stage. As a comparison, previous approaches~\cite{carion2020end, zhu2021deformable, sun2021sparse} rely on the same number of processed boxes during training and evaluation, and their detection decoders are used only once in the forward pass.

\subsection{Discussion}

We conduct a comparative analysis between \ours and previous multi-stage detectors~\cite{ren2015faster,cai2019cascade,carion2020end, sun2021sparse}. Cascade R-CNN adopts a three-stage prediction refinement process where the three stages do not share parameters and are used only once as a complete \emph{head} during the inference phase. Recent works~\cite{carion2020end, zhu2021deformable, sun2021sparse} have adopted a similar structure as Cascade R-CNN but with more stages (\ie, six), following the default setting of DETR~\cite{carion2020end}. While \ours also employs the six-stage structure within its head, the distinguishing feature is that \ours can reuse the entire head multiple times to achieve further performance gains. However, prior works could not improve performance by reusing the detection head in most cases or could only achieve limited performance gains. More detailed results are in \Cref{sec:exp_ablation}.
\section{Experiments}\label{sec:exp}
We first show the attractive flexibility of \ours. Then we compare \ours with previous well-established detectors on COCO~\cite{lin2014microsoft} and CrowdHuman~\cite{shao2018crowdhuman} dataset. Finally, we provide ablation studies on the components of \ours.

\paragraph{COCO}~\cite{lin2014microsoft} dataset contains about 118K training images in the \texttt{train2017} set and 5K validation images in the \texttt{val2017} set. There are 80 object categories in total.  We report box average precision over multiple IoU thresholds~(AP), threshold 0.5~(AP$_{50}$) and 0.75~(AP$_{75}$).

\paragraph{LVIS v1.0}~\cite{gupta2019lvis} dataset is a large-vocabulary object detection and instance segmentation dataset which has 100K training images and 20K validation images. LVIS shares the same source images as COCO, while its annotations capture the long-tailed distribution in 1203 categories. We adopt MS-COCO style box metric AP, AP$_{50}$ and AP$_{75}$ in LVIS evaluation. For LVIS, the training schedule is 210K, 250K, and 270K.
 
\paragraph{CrowdHuman}~\cite{shao2018crowdhuman} dataset is a large dataset covering various crowd scenarios. It has 15K training images and 4.4K validation images, including a total of 470K human instances and 22.6 persons per image. 
Following previous settings~\cite{zhang2017citypersons, lin2021detr, zhu2020cvpods, sun2021sparse}, we adopt evaluation metrics as AP under IoU threshold 0.5. 

\subsection{Implementation Details.}

The ResNet and Swin backbone are initialized with pre-trained weights on ImageNet-1K and ImageNet-21K~\cite{deng2009imagenet}, respectively. The newly added detection decoder is initialized with Xavier init~\cite{glorot2010understanding}. We train \ours using AdamW~\cite{loshchilov2018decoupled} optimizer with the initial learning rate as $2.5\times 10^{-5}$ and the weight decay as $10^{-4}$. All models are trained with a mini-batch size 16 on 8 GPUs. The default training schedule is 450K iterations, with the learning rate divided by 10 at 350K and 420K iterations. 
Data augmentation strategies contain random horizontal flip, scale jitter of resizing the input images such that the shortest side is at least 480 and at most 800 pixels while the longest is at most 1333~\cite{wu2019detectron2}, and random crop augmentations. We do not use the EMA and some strong data augmentation like MixUp~\cite{zhang2018mixup} or Mosaic~\cite{ge2021yolox}.

At the inference stage, we report performances of \ours under diverse settings, which are combinations of different numbers of random boxes and iteration steps. The predictions at each sampling step are ensembled together by NMS to get the final predictions.

\begin{figure}[t]
\begin{subfigure}{0.48\textwidth}
\hspace{-3mm}
\includegraphics[width=1.05\linewidth]{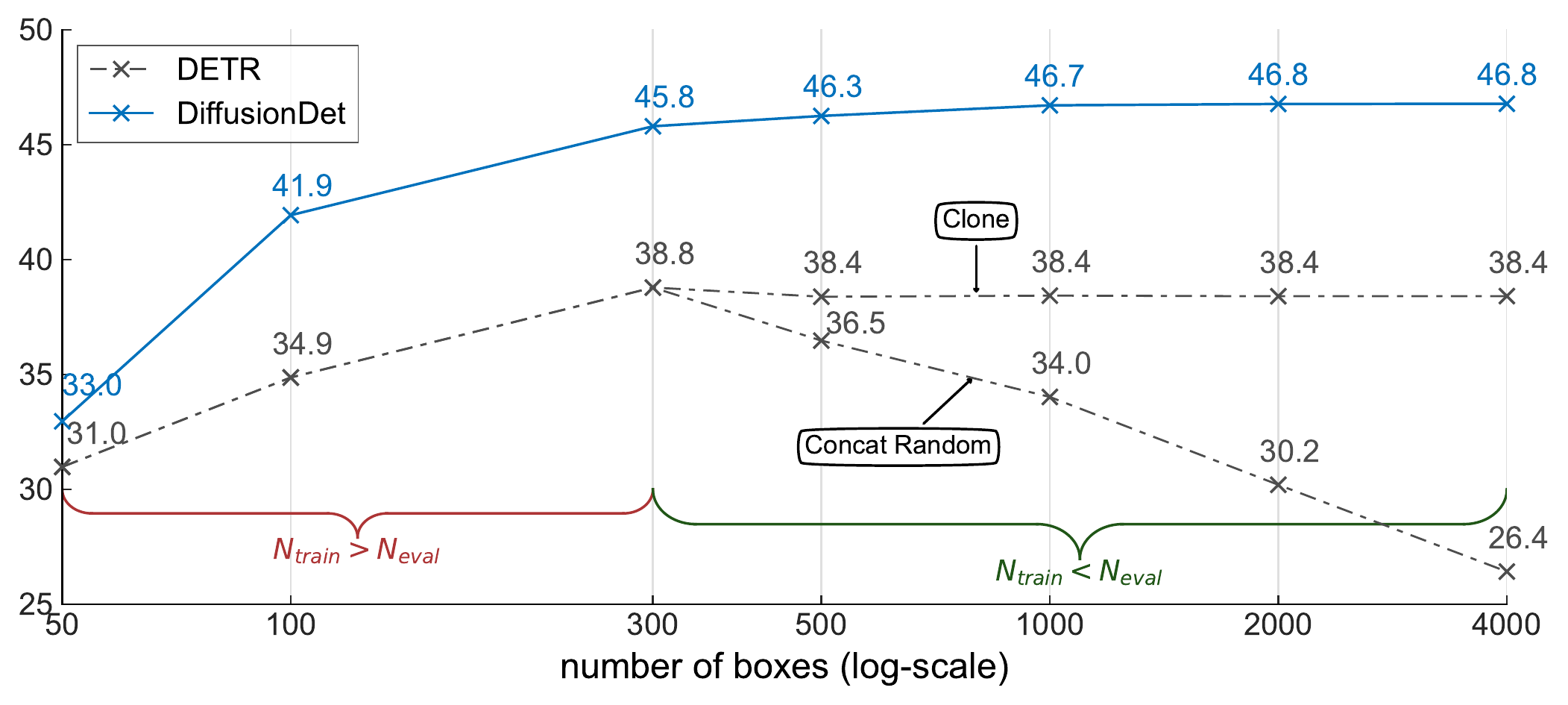}
\caption{\textbf{Dynamic number boxes.} Both DETR and \ours are trained with 300 object queries or proposal boxes. More proposal boxes in inference bring accuracy improvement on \ours, while degenerate DETR.}
\label{fig:eval_boxes}
\vspace{5mm}
\end{subfigure}
\begin{subfigure}{0.48\textwidth}
\hspace{-3mm}
\includegraphics[width=1.05\linewidth]{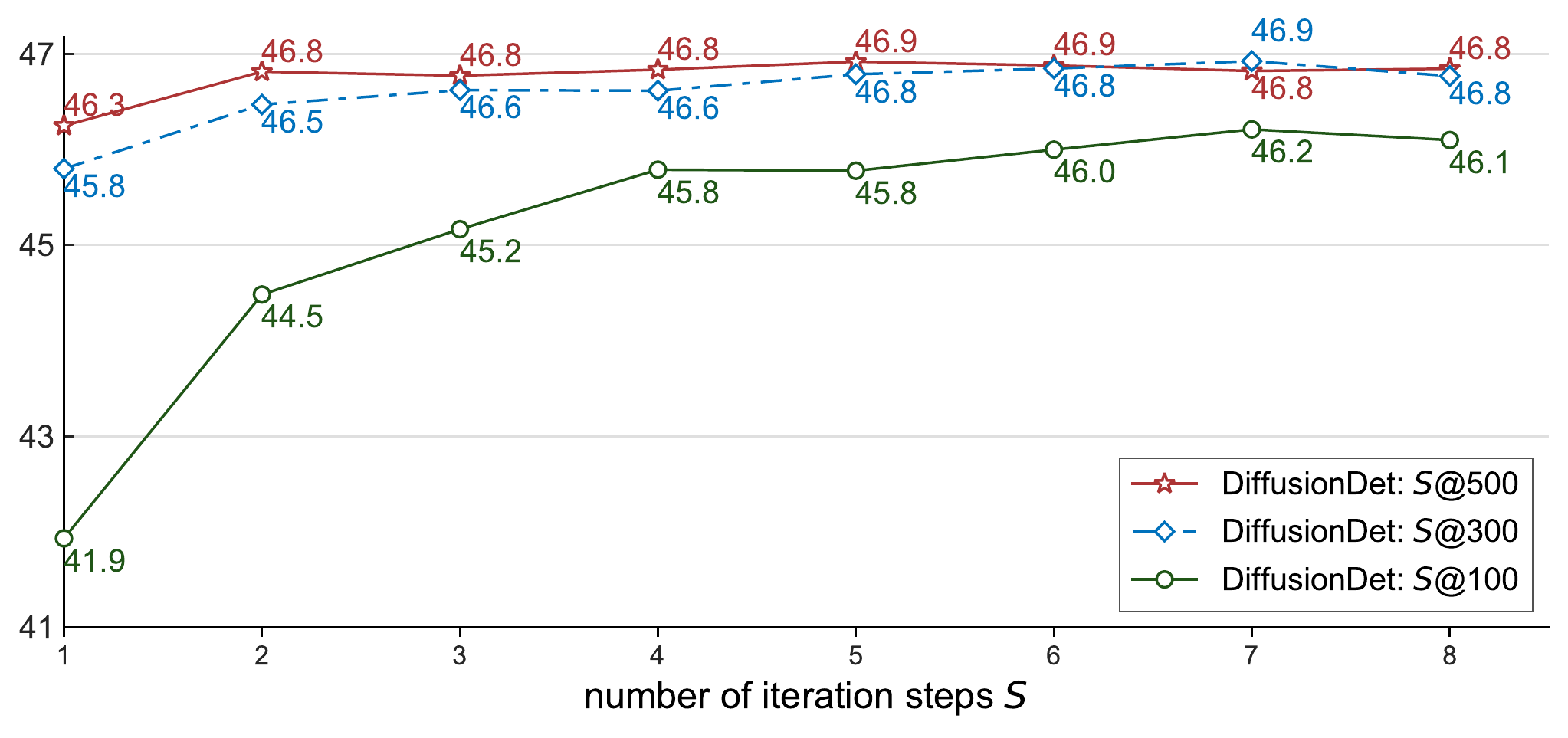}
\caption{\textbf{Iterative evaluation.} `$S$@500' denotes that we evaluate \ours with 500 boxes using a different number of iteration steps. For all cases, the accuracy increases with refinement times.}
\label{fig:sample_steps}
\end{subfigure}
\caption{\textbf{Flexibility of \ours}. All experiments are trained on COCO 2017 \texttt{train} set and evaluated on COCO 2017 \texttt{val} set. \ours uses the same network parameters for all settings in Figure~\ref{fig:eval_boxes} and \ref{fig:sample_steps}. Our proposed \ours is able to benefit from more proposal boxes and iteration steps using the same network parameters.}
\label{fig:once_for_all}
\end{figure}

\subsection{Main Properties}\label{sec:main_pro}

The main properties of \ours lie on \emph{once training for all inference cases}. Once the model is trained, it can be used with changing the number of boxes and the number of iteration steps in inference, as shown in Figure~\ref{fig:once_for_all} and \Cref{tab:zero-shot-crowdhuman}. 
Therefore, we can deploy a single \ours to multiple scenarios and obtain a desired speed-accuracy trade-off without re-training the network.

\paragraph{Dynamic number of boxes.} We compare \ours with DETR~\cite{carion2020end} to show the advantage of dynamic boxes. Comparisons with other detectors are in Appendix B. We reproduce DETR~\cite{carion2020end} with 300 object queries using the official code and default settings for 300 epochs of training. 
We train \ours with 300 random boxes such that the number of candidates is consistent with DETR for a fair comparison.
The evaluation is on \{50, 100, 300, 500, 1000, 2000, 4000\} queries or boxes. 

Since the learnable queries are fixed after training in the original setting of DETR, we propose a simple workaround to enable DETR work with a different number of queries: when $N_{eval} < N_{train}$, we directly choose $N_{eval}$ queries from $N_{train}$ queries; when $N_{eval} > N_{train}$, we clone existing $N_{train}$ queries up to $N_{eval}$~(a.k.a. \texttt{clone}). We equip DETR with NMS because cloned queries will produce similar detection results as the original queries. 
As shown in \Cref{fig:eval_boxes}, the performance of \ours increases steadily with the number of boxes used for evaluation. For example, \ours can achieve 1.0 AP gain when the number of boxes increases from 300 to 4000. On the contrary, cloning more queries for DETR~($N_{eval}>$~300) causes a slight decrease in DETR performance from 38.8 to 38.4 AP, which is then held constant when using more queries.

We also implement another method for DETR when $N_{eval} > N_{train}$, concatenating extra $N_{eval} - N_{train}$ randomly initialized queries~(a.k.a. \texttt{concat random}). With this strategy, DETR has a clear performance drop when the $N_{eval}$ is different from $N_{train}$. Besides, this performance drop becomes larger when the difference between $N_{eval}$ and $N_{train}$ increases. For example, when the number of boxes increases to 4000, DETR only has 26.4 AP with \texttt{concat random} strategy, which is 12.4 lower than the peak value~(\ie, 38.8 AP with 300 queries).

\paragraph{Iterative evaluation.}
We further investigate the performance of our proposed approach by increasing the number of iterative steps from 1 to 8, and the corresponding results are illustrated in \Cref{fig:sample_steps}. Our findings indicate that the \ours models employing 100, 300, and 500 random boxes exhibit consistent performance improvements as the number of iterations increases. Moreover, we observe that \ours with fewer random boxes tends to achieve more substantial gains with refinement. For instance, the AP of \ours instance utilizing 100 random boxes improves from 41.9~(1 step) to 46.1~(8 steps), representing an absolute improvement of 4.2 AP.

\paragraph{Zero-shot transferring.} 
To further validate the effectiveness of generalization, we conduct an evaluation of COCO-pretrained models on the CrowdHuman dataset, without any additional fine-tuning. Specifically, our focus is on the \texttt{[person]} class for the final average precision (AP) performance. The experimental results are presented in \Cref{tab:zero-shot-crowdhuman}.
Our observations indicate that when transferring to a new dataset with scenarios that are more densely populated than COCO, our proposed method, namely \ours, demonstrates a notable advantage by increasing the number of evaluation boxes or iteration steps.
For instance, by increasing the number of boxes from 300 to 2000 and the iteration steps from 1 to 4, \ours achieves a significant AP gain of \textbf{5.3} and \textbf{4.8}, respectively. In contrast, previous methods exhibit limited gain or serious performance degradation, with a decrease of 14.0 AP.
The impressive flexibility of DiffusionDet implies that it is an invaluable asset for object detection tasks across a wide range of scenarios, including sparsely populated and densely crowded environments, without any additional fine-tuning requirements.

\begin{table}[t]
\tablestyle{3.7pt}{1.1}
\begin{tabular}{@{}lcccccc@{}}
\shline
Method & AP & AP$_{50}$ & AP$_{75}$ & AP$_s$ & AP$_m$ & AP$_l$ \\
\shline
\multicolumn{7}{c}{\scriptsize{ResNet-50~\cite{he2016deep}}} \\
RetinaNet~\cite{wu2019detectron2}   & 38.7 & 58.0 & 41.5 & 23.3 & 42.3 & 50.3  \\
Faster R-CNN~\cite{wu2019detectron2} & 40.2 & 61.0 & 43.8 & 24.2 & 43.5 & 52.0  \\
Cascade R-CNN~\cite{wu2019detectron2} & 44.3 &62.2 & 48.0 & 26.6 & 47.7 & 57.7  \\
DETR~\cite{carion2020end}   & 42.0 & 62.4 & 44.2 & 20.5 & 45.8 & 61.1  \\
Deformable DETR~\cite{zhu2021deformable}  & 43.8 & 62.6 & 47.7 & 26.4 & 47.1 & 58.0  \\
Sparse R-CNN~\cite{sun2021sparse}   &  45.0 & 63.4 & 48.2 &26.9 & 47.2 & 59.5  \\
\diffcell{\ours~(1 @ 300)} & \diffcell{45.8} & \diffcell{64.1} & \diffcell{50.4} & \diffcell{27.6} & \diffcell{48.7} & \diffcell{62.2}  \\
\diffcell{\ours~(4 @ 300)}  & \diffcell{46.6} & \diffcell{65.1} & \diffcell{51.3} & \diffcell{28.9} & \diffcell{49.2} & \diffcell{62.1}  \\
\diffcell{\ours~(1 @ 500)}  & \diffcell{46.3} & \diffcell{64.8} & \diffcell{50.7} & \diffcell{28.6} & \diffcell{49.0} & \diffcell{62.1} \\
\diffcell{\ours~(4 @ 500)}  & \diffcell{46.8} & \diffcell{65.3} & \diffcell{51.8} & \diffcell{29.6} & \diffcell{49.3} & \diffcell{62.2} \\
\Xhline{2.\arrayrulewidth}
\multicolumn{7}{c}{\scriptsize{ResNet-101~\cite{he2016deep}}} \\
RetinaNet~\cite{wu2019detectron2}   & 40.4 & 60.2 & 43.2 & 24.0 & 44.3 & 52.2 \\
Faster R-CNN~\cite{wu2019detectron2}    & 42.0 & 62.5 & 45.9 & 25.2 & 45.6 & 54.6 \\
Cascade R-CNN~\cite{mmdetection} & 45.5 & 63.7 & 49.9 & 27.6 & 49.2 & 59.1 \\
DETR~\cite{carion2020end}  & 43.5 & 63.8 & 46.4 & 21.9 & 48.0 & 61.8  \\
Sparse R-CNN~\cite{sun2021sparse}  & 46.4 & 64.6 & 49.5 & 28.3 & 48.3 & 61.6 \\
\diffcell{\ours~(1 @ 300)} & \diffcell{46.7} & \diffcell{65.0} & \diffcell{51.0} & \diffcell{29.6} & \diffcell{49.7} & \diffcell{63.2} \\
\diffcell{\ours~(4 @ 300)} & \diffcell{47.4} & \diffcell{65.8} & \diffcell{52.0} & \diffcell{30.1} & \diffcell{50.4} & \diffcell{63.1} \\
\diffcell{\ours~(1 @ 500)} & \diffcell{47.2} & \diffcell{65.7} & \diffcell{51.6} & \diffcell{30.2} & \diffcell{50.2} & \diffcell{62.7} \\
\diffcell{\ours~(4 @ 500)} &  \diffcell{47.5} & \diffcell{65.7} & \diffcell{52.0} & \diffcell{30.8} & \diffcell{50.4} & \diffcell{63.1} \\
\Xhline{2.\arrayrulewidth}
\multicolumn{7}{c}{\scriptsize{Swin-Base~\cite{liu2021swin}}} \\
Cascade R-CNN~\cite{liu2021swin} & 51.9 & 70.9 & 56.5 & 35.4 & 55.2 & 67.4 \\
Sparse R-CNN & 52.0 & 72.2 &  57.0 & 35.8 & 55.1 & 68.2 \\
\diffcell{\ours~(1 @ 300)} & \diffcell{52.5} & \diffcell{71.8} & \diffcell{57.3} & \diffcell{35.0} & \diffcell{56.4} & \diffcell{69.3} \\
\diffcell{\ours~(4 @ 300)} & \diffcell{53.3} & \diffcell{72.8} & \diffcell{58.6} & \diffcell{36.6} & \diffcell{57.0} & \diffcell{69.2} \\
\diffcell{\ours~(1 @ 500)} & \diffcell{53.0} & \diffcell{72.3} & \diffcell{58.0} & \diffcell{35.5} & \diffcell{56.9} & \diffcell{69.1} \\
\diffcell{\ours~(4 @ 500)} & \diffcell{53.3} & \diffcell{72.7} & \diffcell{58.4} & \diffcell{36.2} & \diffcell{56.9} & \diffcell{69.0}  \\
\shline
\end{tabular}

\caption{\textbf{Comparisons with different object detectors on COCO 2017 \texttt{val} set}. \texttt{[S@$N_{eval}$]} denotes the number of iteration steps \texttt{S} and number of evaluation boxes $N_{eval}$.
The reference after each method indicates the source of its results. The method without reference is our implementation.
}
\label{tab:coco-sota-val}
\vspace{-5pt}
\end{table}

\begin{table}[t]
\tablestyle{4.5pt}{1.1}
\begin{tabular}{@{}lcccccc@{}}
\shline
Method & AP & AP$_{50}$ & AP$_{75}$ & AP$_r$ & AP$_c$ & AP$_f$ \\
\shline
\multicolumn{7}{c}{\scriptsize{ResNet-50~\cite{he2016deep}}} \\
\gc{Faster R-CNN$^\dagger$} & \gc{22.5} & \gc{37.1} & \gc{23.6} & \gc{9.9} & \gc{21.1} & \gc{29.7} \\
\gc{Cascade R-CNN$^\dagger$} & \gc{26.3} & \gc{37.8} & \gc{27.8} & \gc{12.3} & \gc{24.9} & \gc{34.1} \\
Faster R-CNN & 25.2 & 40.6 & 26.9 & 16.4 & 23.4 & 31.1 \\
Cascade R-CNN & 29.4 & 41.4 & 30.9 & 20.0 & 27.7 & 35.4 \\
Sparse R-CNN & 29.2 & 41.0 & 30.7 & 20.6 & 27.7 & 34.6 \\
\diffcell{\ours (1 @ 300)} & \diffcell{29.4} & \diffcell{40.4} & \diffcell{31.0} & \diffcell{22.7} & \diffcell{27.2} & \diffcell{34.7} \\
\diffcell{\ours (1 @ 500)} & \diffcell{30.5} & \diffcell{42.1} & \diffcell{32.1} & \diffcell{23.3} & \diffcell{28.1} & \diffcell{36.3} \\
\diffcell{\ours (1 @ 1000)} & \diffcell{31.4} & \diffcell{43.2} & \diffcell{33.3} & \diffcell{24.5} & \diffcell{28.8} & \diffcell{37.3} \\
\diffcell{\ours (4 @ 300)} & \diffcell{31.5} & \diffcell{43.4} & \diffcell{33.5} & \diffcell{24.1} & \diffcell{29.3} & \diffcell{37.4} \\
\shline
\multicolumn{7}{c}{\scriptsize{ResNet-101~\cite{he2016deep}}} \\
\gc{Faster R-CNN$^\dagger$} & \gc{24.8} & \gc{39.8} & \gc{26.1} & \gc{13.7} & \gc{23.1} & \gc{31.5} \\
\gc{Cascade R-CNN$^\dagger$} & \gc{28.6} & \gc{40.1} & \gc{30.1} & \gc{15.3} & \gc{27.3} & \gc{35.9} \\
Faster R-CNN & 27.2 & 42.9 & 29.1 & 18.8 & 25.4 & 33.0 \\
Cascade R-CNN & 31.6 & 43.8 & 33.4 & 23.9 & 29.8 & 37.0 \\
Sparse R-CNN & 30.1 & 42.0 & 31.9 & 23.5 & 27.5 & 35.9 \\
\diffcell{\ours (1 @ 300)} & \diffcell{30.9} & \diffcell{42.1} & \diffcell{32.6} & \diffcell{22.4} & \diffcell{29.9} & \diffcell{35.8} \\
\diffcell{\ours (1 @ 500)} & \diffcell{31.8} & \diffcell{43.7} & \diffcell{33.6} & \diffcell{23.5} & \diffcell{30.2} & \diffcell{37.3} \\
\diffcell{\ours (1 @ 1000)} & \diffcell{33.0} & \diffcell{45.0} & \diffcell{34.9} & \diffcell{24.8} & \diffcell{31.4} & \diffcell{38.3} \\
\diffcell{\ours (4 @ 300)} & \diffcell{33.0} & \diffcell{45.2} & \diffcell{35.1} & \diffcell{24.2} & \diffcell{31.5} & \diffcell{38.5} \\
\shline
\multicolumn{7}{c}{\scriptsize{Swin-Base~\cite{liu2021swin}}} \\
\diffcell{\ours (1 @ 300)} & \diffcell{39.5} & \diffcell{52.3} & \diffcell{42.0} & \diffcell{33.0} & \diffcell{38.5} & \diffcell{43.5} \\
\diffcell{\ours (1 @ 500)} & \diffcell{40.8} & \diffcell{54.2} & \diffcell{43.6} & \diffcell{33.4} & \diffcell{39.9} & \diffcell{45.2} \\
\diffcell{\ours (1 @ 1000)} & \diffcell{41.9} & \diffcell{55.7} & \diffcell{44.8} & \diffcell{35.3} & \diffcell{40.6} & \diffcell{46.2} \\
\diffcell{\ours (4 @ 300)} & \diffcell{42.0} & \diffcell{55.8} & \diffcell{44.9} & \diffcell{34.8} & \diffcell{40.9} & \diffcell{46.4} \\
\shline
\end{tabular}

\caption{\textbf{Comparisons with different object detectors on LVIS v1.0 \texttt{val} set}. We re-implement all detectors using federated loss~\cite{zhou2021probablistic} except for the rows in light gray~(with $^\dagger$).
}
\label{tab:lvis}
\end{table}

\begin{table*}[t]
\centering
\newcommand\tr{\rule{0pt}{10pt}}
\newcommand\br{\rule[-3.3pt]{0pt}{3.3pt}}
%#################################################
% Signal Scale
%#################################################
\subfloat[
\textbf{Signal scale}. A large scaling factor can improve detection performance.
\label{tab:signal_scale}
]{
\centering
\begin{minipage}{0.25\linewidth}{\begin{center}
\tablestyle{2pt}{1.05}
\begin{tabular}{x{18}x{24}x{24}x{24}}
\shline
scale & AP & AP$_{50}$ & AP$_{75}$ \\
\shline
0.1 & 38.9 & 54.3 & 42.1 \\
1.0 & 45.0 & 63.0 & 48.9 \\
\bestcell{2.0} & \bestcell{45.8} & \bestcell{64.1} & \bestcell{50.4} \\
3.0 & 45.6 & 63.9 & 50.0 \\
\shline
\end{tabular}
\end{center}}
\end{minipage}
}
\hspace{2em}
%#################################################
% GT Bxoes Padding strategy
%#################################################
\subfloat[
\textbf{GT boxes padding.} Concatenating Gaussian boxes works best.
\label{tab:boxes_padding}
]{
\centering
\begin{minipage}{0.25\linewidth}{\begin{center}
\tablestyle{2pt}{1.05}
\begin{tabular}{y{43}x{24}x{24}x{24}}
\shline
case & AP & AP$_{50}$ & AP$_{75}$ \\
\shline
Repeat    &  44.2 & 62.0 & 48.3 \\
\bestcell{Cat Gaussian} & \bestcell{45.8} & \bestcell{64.1} & \bestcell{50.4} \\
Cat Uniform & 45.2 & 63.3 & 49.3 \\
Cat Full    & 45.7 & 63.9 & 49.9 \\
\shline
\end{tabular}
\end{center}}
\end{minipage}
}
\hspace{2em}
%#################################################
% DDIM 
%#################################################
\subfloat[
\textbf{Sampling strategy.}  Using both DDIM and box renewal works best.
\label{tab:ddim_renewal}
]{
\centering
\begin{minipage}{0.33\linewidth}{\begin{center}
\tablestyle{2pt}{1.05}
\begin{tabular}{x{24}x{42}x{24}x{24}x{24}}
\shline
DDIM & box renewal & iter 1 & iter 2 & iter 3 \\
\shline
           &            & 45.8 & 44.4 & 44.1 \\
\checkmark &            & 45.8 & 46.0 & 46.1 \\
           & \checkmark & 45.8 & 46.3 & 46.3 \\
\bestcell{\checkmark} & \bestcell{\checkmark} & \bestcell{45.8} & \bestcell{46.5} & \bestcell{46.6} \\
\shline
\end{tabular}
\end{center}}
\end{minipage}
}

\caption{\textbf{\ours ablation experiments} on COCO. We report AP, AP$_{50}$, and AP$_{75}$. If not specified, the default setting is:
the backbone is ResNet-50~\cite{he2016deep} with FPN~\cite{lin2017feature}, the signal scale is 2.0, ground-truth boxes padding method is concatenating Gaussian random boxes, DDIM and box renewal are used in sampling step.
Default settings are marked in \colorbox{bestcolor}{gray}.}
\end{table*}

\subsection{Benchmarking on Detection Datasets}\label{sec:benchmark}
In \Cref{tab:coco-sota-val}, we present a comparison of our \ours with several state-of-the-art detectors~\cite{ren2015faster,lin2017focal,cai2019cascade,carion2020end,zhu2021deformable,sun2021sparse} on the COCO dataset. For more comprehensive experimental settings, please refer to the Appendix. Notably, our \ours~(1 @ 300), which adopts a single iteration step and 300 evaluation boxes, achieves an AP of 45.8 with a ResNet-50 backbone, surpassing the performance of several well-established methods such as Faster R-CNN, RetinaNet, DETR, and Sparse R-CNN by a considerable margin. Moreover, \ours can further enhance its superiority by increasing the number of iterations and evaluation boxes.
Besides, \ours shows steady improvement when the backbone size scales up. \ours with ResNet-101~(1 @ 300) achieves 46.7. When using ImageNet-21k pre-trained Swin-Base~\cite{liu2021swin} as the backbone, \ours obtains 52.5 AP, outperforming strong baselines such as Cascade R-CNN and Sparse R-CNN.

Our current model is still lagging behind behind some well developed works like DINO~\cite{zhang2022dino} since it uses some more advanced components such as deformable attention~\cite{zhu2021deformable}, wider detection head. Some of these techniques are orthogonal to DiffusionDet and we will explore to incorporate these to our current pipeline for further improvement.

Experimental results on LVIS are presented in \Cref{tab:lvis}. We reproduce Faster R-CNN and Cascade R-CNN based on \texttt{detectron2}~\cite{wu2019detectron2} while Sparse R-CNN on its original code. 
We first reproduce Faster R-CNN and Cascade R-CNN using the default settings of \texttt{detectron2}, achieving 22.5/24.8 and 26.3/28.8 AP~(with $^\dagger$ in \Cref{tab:lvis}) with ResNet-50/101 backbone, respectively. Further, we boost their performance using the federated loss in~\cite{zhou2021probablistic}. Since images in LVIS are annotated in a federated way~\cite{gupta2019lvis}, the negative categories are sparsely annotated, which deteriorates the training gradients, especially for rare classes~\cite{tan2020equalization}. Federated loss is proposed to mitigate this issue by sampling a subset $S$ of classes for each training image that includes all positive annotations and a random subset of negative ones. Following~\cite{zhou2021probablistic}, we choose $|S| = 50$ in all experiments. Faster R-CNN and Cascade R-CNN earn about 3 AP gains with federated loss. All following comparisons are based on this loss. 

We see that \ours attains remarkable gains using more evaluation steps, with both small and large backbones. 
Moreover, we note that iterative evaluation brings more gains on LVIS compared with COCO. For example, its performance increases from 45.8 to 46.6~(+ 0.8 AP) on COCO while from 29.4 to 31.5~(+2.1 AP) on LVIS, which demonstrates that our \ours would become more helpful for a more challenging benchmark.

\subsection{Ablation Study}\label{sec:exp_ablation}

We conduct ablation experiments on COCO to study \ours in detail. All experiments use ResNet-50 with FPN as the backbone and 300 random boxes for inference without further specification. 

\paragraph{Signal scaling.}
The signal scaling factor controls the signal-to-noise ratio~(SNR) of the diffusion process. 
We study the influence of scaling factors in \Cref{tab:signal_scale}. Results demonstrate that the scaling factor of 2.0 achieves optimal AP performance, outperforming the standard value of 1.0 in image generation task~\cite{ho2020denoising, chen2022analog} and 0.1 used for panoptic segmentation~\cite{chen2022generalist}. We explain that it is because one box only has four representation parameters, \ie, center coordinates~$(c_x, c_y)$ and box size~$(w, h)$, which is coarsely analogous to an image with only four pixels in image generation. The box representation is more fragile than the dense representation, \eg, $512\times 512$ mask presentation in panoptic segmentation~\cite{chen2022analog}.
Therefore, \ours prefers an easier training objective with an increased signal-to-noise ratio compared to image generation and panoptic segmentation.

\paragraph{GT boxes padding strategy.} As introduced in \Cref{sec:method_train}, we need to \emph{pad} additional boxes to the original ground truth boxes such that each image has the same number of boxes. We study different padding strategies in \Cref{tab:boxes_padding}, including (1) repeating original ground truth boxes evenly until the total number reaches pre-defined value $N_{train}$; (2) padding random boxes that follow Gaussian distribution; (3) padding random boxes that follow uniform distribution; (4) padding boxes that have the same size as the whole image, which is the default initialization of learnable boxes in~\cite{sun2021sparse}. Concatenating Gaussian random boxes works best for \ours. 
We use this padding strategy as default.

\paragraph{Sampling strategy.}
We compare different sampling strategies in \Cref{tab:ddim_renewal}. When evaluating \ours that does not use DDIM, we directly take the output prediction of the current step as input for the next step. We found that the AP of \ours degrades with more iteration steps when neither DDIM nor box renewal is adopted.
Besides, only using DDIM or box renewal would bring slight benefits at 3 iteration steps. 
Moreover, our \ours attains remarkable gains when equipped with both DDIM and renewal. These experiments together verify the necessity of both DDIM and box renewal in the sampling step.

\begin{table}[t]
\renewcommand{\arraystretch}{1.05}
\setlength{\tabcolsep}{7pt}
    \centering
    \small
    \begin{tabular}{c |c c c c c}
    \shline
     \diagbox[height=15 pt]{\small train}{\small eval} & 100 & 300 & 500 & 1000 & 2000 \\
    \shline
    100  & 42.9 & 44.4 & 44.5 & 44.6 & 44.6 \\
    300  & 42.8 & 45.7 & 46.2 & 46.3 & 46.4 \\
    500  & 41.9 & 45.8 & 46.3 & 46.7 & 46.8 \\
    \shline
    \end{tabular}
    \caption{\textbf{Matching between training and inference box numbers} on COCO. \ours decouples the number of boxes during the training and inference stages and works well with flexible combinations.}
    \label{tab:train_eval_boxes}
\end{table}

\paragraph{Matching between $N_{train}$ and $N_{eval}$.}  As discussed in \cref{sec:main_pro}, \ours has an appealing property of evaluating with an arbitrary number of random boxes.
To study how the number of training boxes affects inference performance, we train \ours with $N_{train}\in \{100, 300, 500\}$ random boxes separately and then evaluate each of these models with $N_{eval}\in\{100, 300, 500, 1000, 2000\}$. The results are summarized in \Cref{tab:train_eval_boxes}. First, no matter how many random boxes \ours uses for training, the accuracy increases steadily with the $N_{eval}$ until the saturated point at around 2000 random boxes. Second, \ours tends to perform better when the $N_{train}$ and $N_{eval}$ \emph{matches} with each other. For example, \ours trained with $N_{train} = 100$ boxes behaves better than $N_{train} = 300$ and $500$ when $N_{eval} = 100$.

\begin{table}[t]
    \centering
    \tabcolsep=0.18cm
    \footnotesize
    \begin{tabular}{l c c  l l c }
    \toprule
    Method  & \footnotesize{Train}  & \footnotesize{Test}  & \multicolumn{1}{c}{\footnotesize COCO} & \multicolumn{1}{c}{\footnotesize CrowdHuman} & \multicolumn{1}{c}{FPS} \\
    \midrule
     \multirow{2}{*}{\tabincell{c}{DETR~\cite{carion2020end}}} & 6$\times$1 & 6$\times$1 & 42.0 & 61.3 & 39 \\
     & 6$\times$1 & 6$\times$2  & 41.6~\zzz{-0.4} & 62.5~\bbb{+1.2} &  32 \\
    \midrule
    \multirow{3}{*}{\tabincell{c}{\footnotesize Sparse\\ \footnotesize R-CNN~\cite{sun2021sparse}}} & 6$\times$1 & 6$\times$1 &  45.0 &  66.6 & 30 \\
     & 6$\times$1  & 6$\times$2  & 43.6~\zzz{-1.4}  & 60.6~\zzz{-6.0} & 21 \\
     & 12$\times$1 & 12$\times$1 & 44.7~\zzz{-0.3}  & 66.1~\zzz{-0.5}& 21 \\
    \midrule
    \multirow{2}{*}{\footnotesize \ours}   & 6$\times$1 & 6$\times$1  & 45.8 & 66.6 & 30  \\
     & 6$\times$1 & 6$\times$2 & 46.5~\aaa{+0.7}   & 69.7~\aaa{+3.1}  & 20 \\
     \footnotesize \ours$^\dagger$ & 6$\times$1 & 6$\times$1  & 46.8~\aaa{+1.0} & 71.0~\aaa{+4.4} & 24 \\
    \bottomrule
    \end{tabular}
    \vspace{-5pt}
    \caption{\textbf{Running time \vs performance.} $^\dagger$ denotes \ours with 1000 boxes. \texttt{\#Stages}$\times$\texttt{\#Heads} denotes the number of \texttt{stages} and \texttt{heads} utilized during training and test phases. The definitions of \texttt{Stage} and \texttt{Head} are illustrated in \Cref{fig:head_details}.}
    \label{tab:speed}
    \vspace{-10pt}
\end{table}

\noindent
\textbf{Running time \vs accuracy}.
We investigate the running time of \ours under multiple settings, which are evaluated on a single NVIDIA A100 GPU with a mini-batch size of 1. We utilize the notation \texttt{\#Stages}$\times$\texttt{\#Heads} to indicate the number of \texttt{stages} and \texttt{heads} utilized during the training and test phases, as depicted in \Cref{fig:head_details} and results of our investigation are presented in \Cref{tab:speed}.

First, our findings indicate that \ours with a single iteration step and 300 evaluation boxes demonstrate a comparable speed to Sparse R-CNN, achieving 30 and 31 frames per second (FPS), respectively. \ours also showcases similar zero-shot transfer performance on CrowdHuman while outperforming Sparse R-CNN with an 45.8 AP as opposed to 45.0 AP on COCO. Besides, Sparse R-CNN's utilization of the six stages twice results in a 1.4 AP drop (from 45.0 to 43.6) on COCO and a 6.0 AP drop (from 66.6 to 60.6) on CrowdHuman. Similarly, DETR experiences 0.4 performance drop on COCO but 1.2 performance gain on CrowdHuman.

When increasing the number of iteration steps, \ours achieves a 0.7 AP gain on COCO and a 3.1 AP gain on CrowdHuman. And \ours obtains clear performance gains with 1000 evaluation boxes. However, neither DETR nor Sparse R-CNN can achieve performance gains with additional iteration steps. Even if we expand the number of stages to 12, it 
can cause performance degradation for Sparse R-CNN.

It is worth noting that in this work, we have utilized the most fundamental diffusion strategy, DDIM, in our pioneering exploration of using generation models for perception tasks. Similar to the Diffusion model employed in generation tasks, \ours may suffer from a relatively slow sampling speed. Nonetheless, a series of recent works~\cite{salimans2022progressive, lu2022dpmsolver, dockhorn2022genie, song2023consistency} have been proposed to improve the sampling efficiency of the diffusion model. For instance, the most recent consistency models~\cite{song2023consistency} have proposed a fast one-step generation method for the diffusion model. We believe that a more advanced diffusion strategy could potentially address the issue of decreased speed performance of \ours, which we plan to explore in future work.

\begin{figure}[ht]
    \centering
    \includegraphics[width=0.98\linewidth]{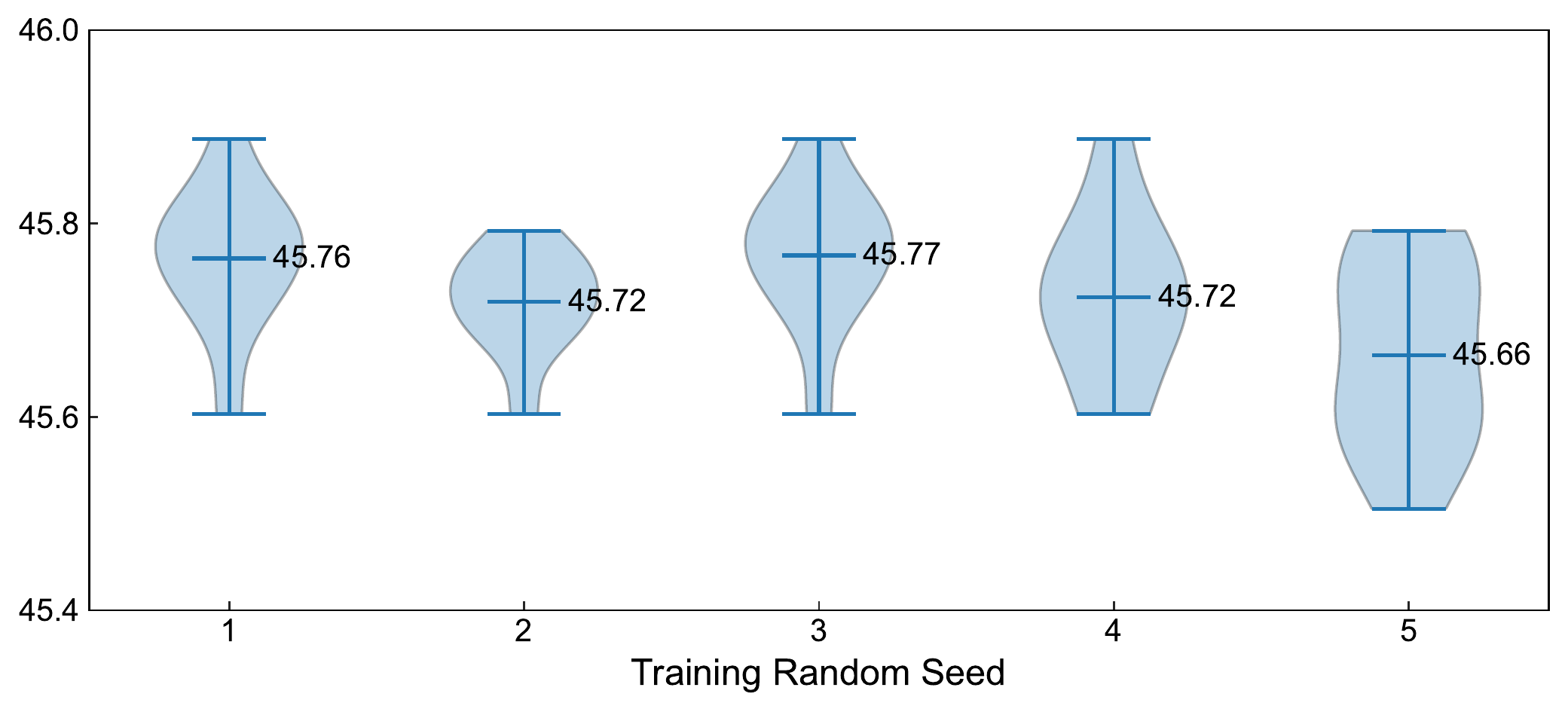}
    \caption{\textbf{Statistical results} over 5 independent training instances, each is evaluated 10 times with different random seeds.}
    \label{fig:violin}
\end{figure}

\paragraph{Random Seed}
Since \ours is given random boxes as input at the start of inference, one may ask whether there is a large performance variance across different random seeds.
We evaluate the stability of \ours by training five models independently with the same configurations except for random seed. Then, we evaluate each model instance with ten different random seeds to measure the distribution of performance, inspired by~\cite{picard2021torch, wightman2021resnet}. 
As shown in \Cref{fig:violin}, most evaluation results are distributed closely to 45.7 AP. Besides, the performance differences among different model instances are marginal, demonstrating that \ours is robust to the random boxes and produces reliable results.

\subsection{Full-tuning on CrowdHuman}
\begin{table}[t]
    \centering
    \small
    \begin{tabular}{l|c c c}
    \shline
    method & AP$_{50}$~$\uparrow$ & mMR~$\downarrow$  & Recall~$\uparrow$ \\
    \shline
    Faster R-CNN~\cite{ren2015faster} & 85.0 & 50.4 & 90.2  \\
    RetinaNet~\cite{lin2017focal}   & 81.7 & 57.6 &  88.6 \\
    FCOS~\cite{tian2019fcos} & 86.1 & 55.2 & 94.3 \\
    DETR~\cite{carion2020end} & 66.1 & 80.6 & - \\
    Deformable DETR~\cite{zhu2021deformable} & 86.7 & 54.0 & 92.5 \\
    Sparse R-CNN~\cite{sun2021sparse} (500)  & 89.2 & 48.3 & 95.9 \\
    Sparse R-CNN~\cite{sun2021sparse} (1000) & 89.7 & 49.1 & 97.5 \\
    \hline
    \ours (1 @ 1000) & 90.1 & 46.5 & 96.2 \\
    \ours (1 @ 3000) & 91.2 & 47.7 & 98.4 \\
    \ours (3 @ 1000) & \bf{91.4} & \bf{45.7} & \bf{98.4} \\
    \shline
    \end{tabular}
    \vspace{-5pt}
    \caption{\textbf{Full tuning on CrowdHuman.} }
    \label{tab:fulltune_crowdhuman}
    \vspace{-10pt}
\end{table}

In addition to the cross-dataset generalization evaluation from COCO to CrowdHuman discussed in \Cref{sec:main_pro}, we further full-tune \ours on CrowdHuman. The comparison results are shown in \Cref{tab:fulltune_crowdhuman}. We see that \ours achieves superior performance compared with previous methods. For example, with a single step and 1000 boxes, \ours obtains 90.1 AP$_{50}$, outperforming Sparse R-CNN with 1000 boxes. Besides, further increasing boxes to 3000 and iteration steps can both bring performance gains.

\section{Conclusion}

In this work, we propose a novel detection paradigm, \ours,  by viewing object detection as a denoising diffusion process from noisy boxes to object boxes. Our noise-to-box pipeline has several appealing properties, including the dynamic number of boxes and iterative evaluation, enabling us to use the same network parameters for flexible evaluation without re-training the model. Experiments on standard detection benchmarks show that \ours achieves favorable performance compared to well-established detectors.  

\vspace{5pt}
\noindent\textbf{Acknowledgement.} This paper is partially supported by the National Key R\&D Program of China No.2022ZD0161000 and the General Research Fund of Hong Kong No.17200622.

\appendix
\setcounter{page}{1}

\section{Formulation of Diffusion Model}\label{appendix:diffusion_formula}

We provide a detailed review of the formulation of diffusion models, following the notion of~\cite{ho2020denoising, dhariwal2021diffusion, nichol2021improved}. Starting from a data distribution $\bm{z}_0 \sim q(\bm{z}_0)$, we define a forward Markovian noising process $q$ which produces data samples $\bm{z}_1$, $\bm{z}_2$, ..., $\bm{z}_T$ by gradually adding Gaussian noise at each timestep $t$. In particular, the added noise is scheduled by the variance $\beta_t \in (0, 1)$:
\begin{alignat}{2}
    q(\bm{z}_{1:T}|\bm{z}_0) &\coloneqq \prod^T_{t=1}q(\bm{z}_t | \bm{z}_{t-1}) \\
    q(\bm{z}_t | \bm{z}_{t-1}) &\coloneqq \mathcal{N}(\bm{z}_t; \sqrt{1-\beta_t}\bm{z}_{t-1}, \beta_t \bm{I})
\end{alignat}

\noindent As noted by Ho~\etal~\cite{ho2020denoising}, we can directly sample data $\bm{z}_t$ at an arbitrary timestep $t$ without the need of applying $q$ repeatedly:

\begin{alignat}{2}
    q(\bm{z}_t|\bm{z}_0) &\coloneqq \mathcal{N}(\bm{z}_t; \sqrt{\bar{\alpha}_t}\bm{z}_0, (1 - \bar{\alpha}_t)\bm{I}) \\
    &\coloneqq \sqrt{\bar{\alpha}_t} \bm{z}_0  + \epsilon \sqrt{1 - \bar{\alpha}_t}, \epsilon \in \mathcal{N}(0, \bm{I})\label{eq:diffuse}
\end{alignat}
\noindent where $\bar{\alpha}_t \coloneqq \prod_{s=0}^{t} \alpha_s$ and $\alpha_t \coloneqq 1 - \beta_t$. Then, we could use $\bar{\alpha}_t$ instead of $\beta_t$ to define the noise schedule.

Based on Bayes' theorem, it is found that the posterior $q(\bm{z}_{t-1}|\bm{z}_t, \bm{z}_0)$ is a Gaussian distribution as well:

\begin{alignat}{2}
     q(\bm{z}_{t-1}|\bm{z}_t, \bm{z}_0) &= \mathcal{N}(\bm{z}_{t-1}; \tilde{\mu}(\bm{z}_t, \bm{z}_0), \tilde{\beta}_t \mathbf{I})\label{eq:posterior}
\end{alignat}

\noindent where
\begin{alignat}{2}
    \tilde{\mu}_t(\bm{z}_t, \bm{z}_0) &\coloneqq
    \frac{\sqrt{\bar{\alpha}_{t-1}}\beta_t}{1-\bar{\alpha}_t}\bm{z}_0 + \frac{\sqrt{\alpha_t}(1-\bar{\alpha}_{t-1})}{1-\bar{\alpha}_t} \bm{z}_t \label{eq:mutilde}
\end{alignat}
\noindent and
\begin{alignat}{2}
    \tilde{\beta}_t &\coloneqq \frac{1-\bar{\alpha}_{t-1}}{1-\bar{\alpha}_t} \beta_t \label{eq:betatilde}
\end{alignat}
 
\noindent are the mean and variance of this Gaussian distribution.

We could get a sample from $q(\bm{z}_0)$ by first sampling from $q(\bm{z}_T)$ and running the reversing steps $q(\bm{z}_{t-1} | \bm{z}_t)$ until $\bm{z}_0$. Besides, the distribution of $q(\bm{z}_T)$ is nearly an isotropic Gaussian distribution with a sufficiently large $T$ and reasonable schedule of $\beta_t$~($\beta_t \rightarrow 0$), which making it trivial to sample $\bm{z}_T \sim \mathcal{N}(0, \bm{I})$. Moreover, since calculating $q(\bm{z}_{t-1} | \bm{z}_t)$ exactly should depend on the entire data distribution, we could approximate $q(\bm{z}_{t-1} | \bm{z}_t)$ using a neural network,
which is optimized to predict a mean $\mu_\theta$ and a diagonal covariance matrix $\Sigma_\theta$:
\begin{alignat}{2}
p_{\theta}(\bm{z}_{t-1}|\bm{z}_t) &\coloneqq \mathcal{N}(\bm{z}_{t-1};\mu_{\theta}(\bm{z}_t, t), \Sigma_{\theta}(\bm{z}_t, t)) \label{eq:ptheta}
\end{alignat}
Instead of directly parameterizing $\mu_\theta(\bm{z}_t, t)$, Ho~\etal~\cite{ho2020denoising} found learning a network $f_\theta(\bm{z}_t, t)$ to predict the $\epsilon$ or $\bm{z}_0$ from \Cref{eq:diffuse} worked best. We choose to predict $\bm{z}_0$ in this work.

\section{Additional Experiments}\label{sec:app_exps}

We provide some additional experiments in this section for more detailed analysis.

\subsection{Dynamic Number of Boxes}

\begin{figure}[t]
    \centering
    \begin{subfigure}{\linewidth}
    \includegraphics[width=0.98\linewidth]{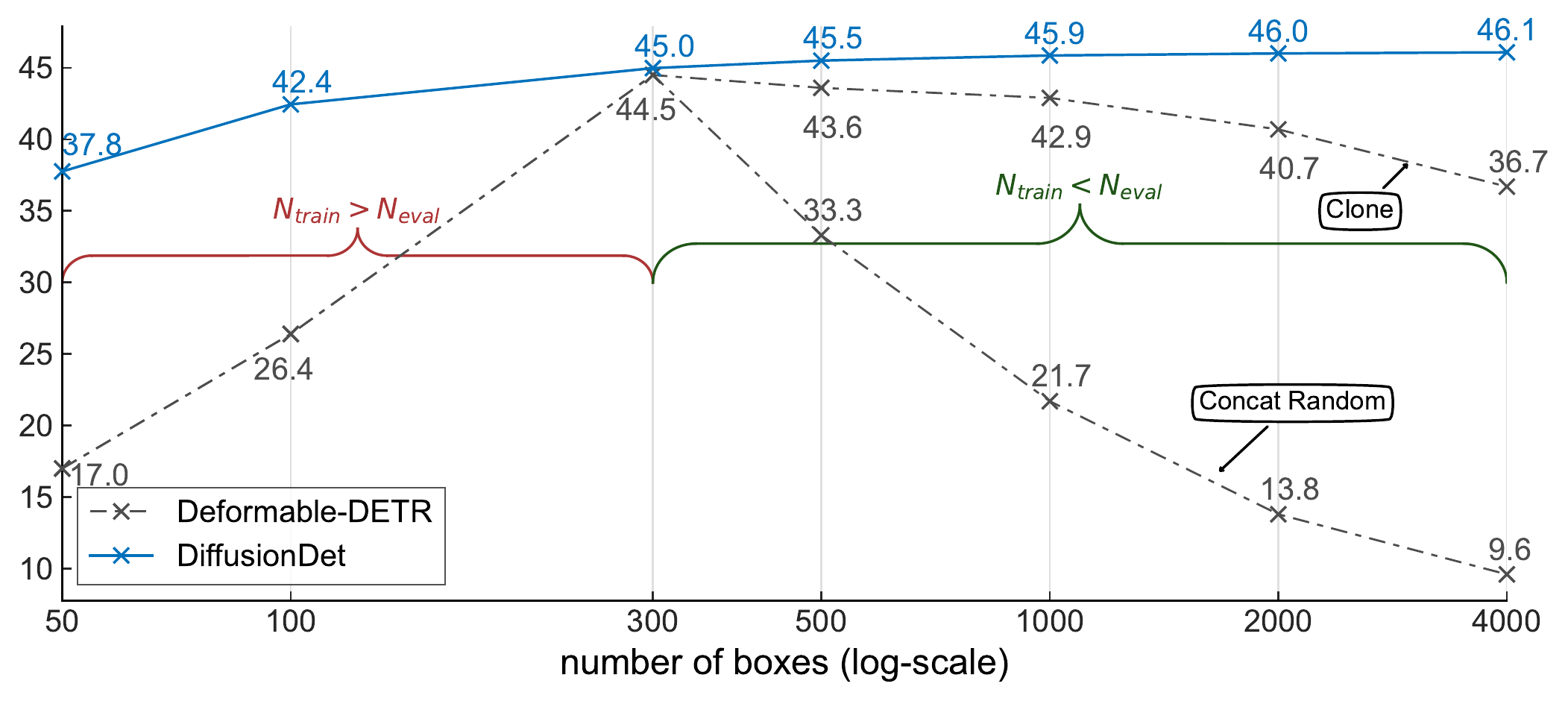}
    \caption{\textbf{Deformable DETR~\cite{zhu2021deformable} vs. \ours}}
    \end{subfigure}
    \begin{subfigure}{\linewidth}
    \includegraphics[width=0.98\linewidth]{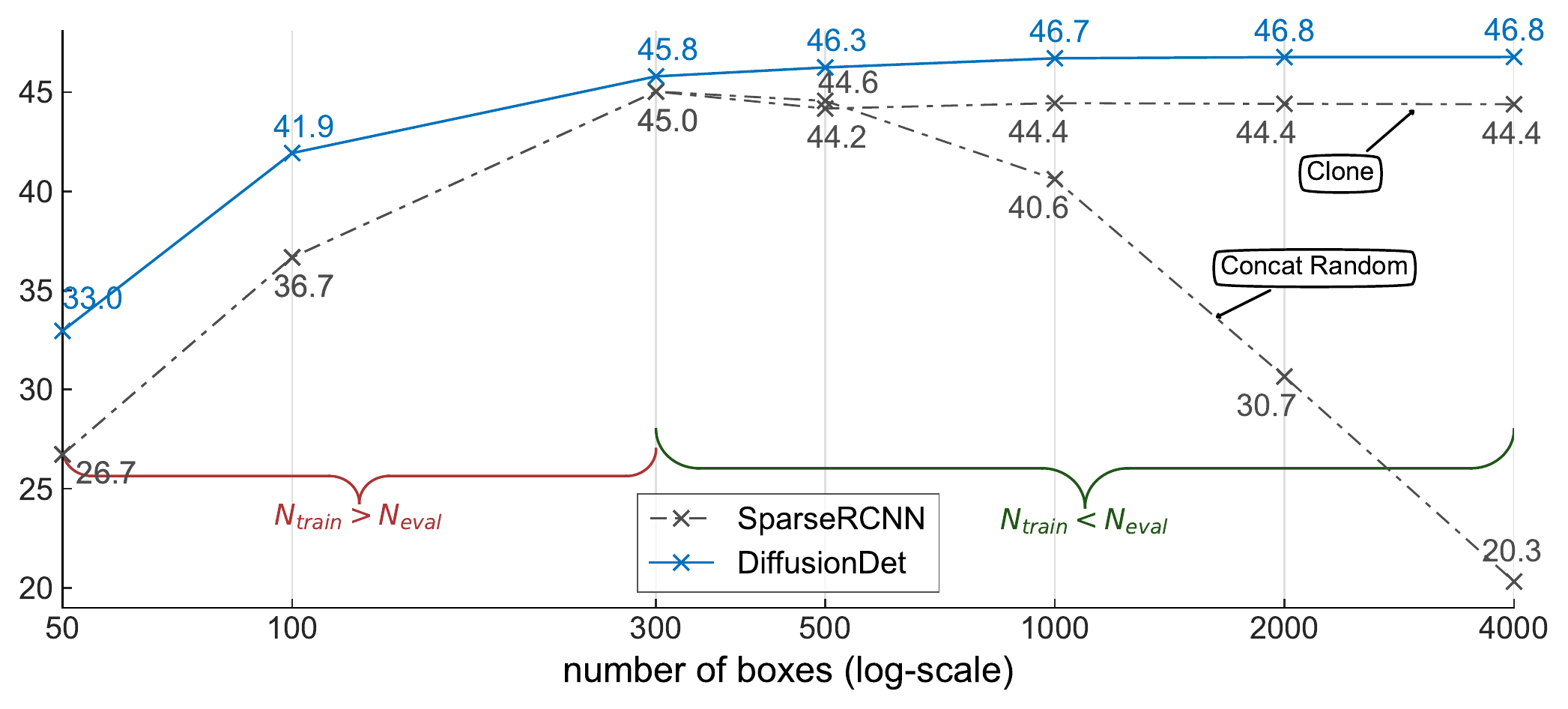}
    \caption{\textbf{Sparse R-CNN~\cite{sun2021sparse} vs. \ours}}
    \end{subfigure}
    \caption{\textbf{Dynamic number of boxes.} All models are trained with 300 candidates~(\ie, learnable queries or random boxes). When $N_{train} > N_{eval}$, we directly choose $N_{eval}$ from $N_{train}$ candidates; when $N_{train} < N_{eval}$, we design two strategies, \ie, \texttt{clone} and \texttt{concat random}.}
    \label{fig:num_boxesv2}
\end{figure}

We further compare the dynamic box property of \ours with Deformable DETR~\cite{zhu2021deformable} and Sparse R-CNN~\cite{sun2021sparse} in \Cref{fig:num_boxesv2}. We directly use the provided models in their official code repositories.\footnote{\scriptsize\url{https://github.com/fundamentalvision/Deformable-DETR}}
\footnote{\scriptsize\url{https://github.com/PeizeSun/SparseR-CNN}}

We make Deformable DETR work under $N_{train} \neq N_{eval}$ setting using the same \texttt{clone} and \texttt{concat random} strategies as DETR as introduced in \Cref{sec:main_pro}. For Sparse R-CNN, the strategy \texttt{concat random} is slightly different since Sparse R-CNN has both learnable queries and learnable boxes. Therefore, we concatenate $N_{eval} - N_{train}$ boxes to existing $N_{train}$ boxes which are initialized to have the same size as the whole image. Besides, we also concatenate $N_{eval} - N_{train}$ randomly initialized queries to existing $N_{train}$ queries in the same way as DETR and Deformable DETR.

Similar to DETR~\cite{carion2020end}, neither Deformable DETR nor Sparse R-CNN has the dynamic box property.  
Specifically, the performance of Deformable DETR decreases to 9.6 AP when $N_{eval} = 4000$, far lower than the peak value of 44.5 AP. Although Sparse R-CNN has a slower decrease compared with Deformable DETR, its performance is unsatisfactory when the $N_{eval}$ is inconsistent with $N_{train}$. These findings suggest the distinctive dynamic number of boxes property of \ours.

\subsection{Iterative Evaluation}\label{sec:app_refine}

\begin{table}[t]
\tablestyle{3.2pt}{1.3}
    \centering
    \small
    \definecolor{mygreen}{RGB}{83,161,81}
\definecolor{negcolor}{RGB}{103,81,165}
\newcommand{\posacc}[1]{{\scriptsize\selectfont \color{mygreen}~(+#1)}}
\newcommand{\negacc}[1]{{\scriptsize\selectfont \color{negcolor}~(-#1)}}

\begin{tabular}{l | c | c c c}
    \shline
    method & \texttt{[E]} & step 1 & step 3 & step 5 \\
    \shline
    \multirow{2}{*}{DETR} & & \multirow{2}{*}{42.03} & 42.00\negacc{0.03} & 41.88\negacc{0.15} \\
    & \checkmark &   & 41.35\negacc{0.68} & 41.36\negacc{0.67} \\
    \hline
    \multirow{2}{*}{Deformable DETR} & & \multirow{2}{*}{44.46} & 43.45\negacc{1.01}  & 43.40\negacc{1.06} \\
       &\checkmark&    & 44.03\negacc{0.43}  &  44.04\negacc{0.42} \\
     \hline
     \multirow{2}{*}{Sparse R-CNN} & & \multirow{2}{*}{45.02} & 1.32\negacc{43.70} &  0.32\negacc{44.70}\\
                                             & \checkmark & & 42.90\negacc{2.12} & 42.24\negacc{2.78} \\
     \hline
     \multirow{2}{*}{\ours} & & \multirow{2}{*}{45.80} & 45.74\negacc{0.06} & 45.46\negacc{0.34} \\
     &  \checkmark &   & 46.62\posacc{0.82} & 46.79\posacc{0.99} \\
     \shline
    \end{tabular}
    \vspace{-10pt}
    \caption{\textbf{Iterative evaluation.} \texttt{[E]} denotes ensembling predictions from multiple steps. NMS is adopted when using an ensemble strategy. We show the performance differences of each method with respect to their own performance on step 1 by \negacc{} or \posacc{}.}
    \label{tab:ensemble}
    \vspace{-14pt}
\end{table}

In \Cref{tab:ensemble} we compare the progressive refinement property of \ours with some previous approaches like DETR~\cite{carion2020end}, Deformable DETR~\cite{zhu2021deformable} and Sparse R-CNN~\cite{sun2021sparse}. All of these four models have 6 cascading stages as the detection decoder. The refinement refers to the output of the previous 6 stages and is taken as the input of the next 6 stages. 
All model checkpoints are from Model Zoo in their official code repositories.

We experiment with two settings: (1) only use the output of the last reference step as the final prediction; (2) use the ensemble of the output of multiple steps as the final prediction. For the latter setting, we adopt NMS to remove duplicate predictions among different steps.

\begin{figure}[t]
    \centering
    \begin{subfigure}{\linewidth}
    \includegraphics[width=0.98\linewidth]{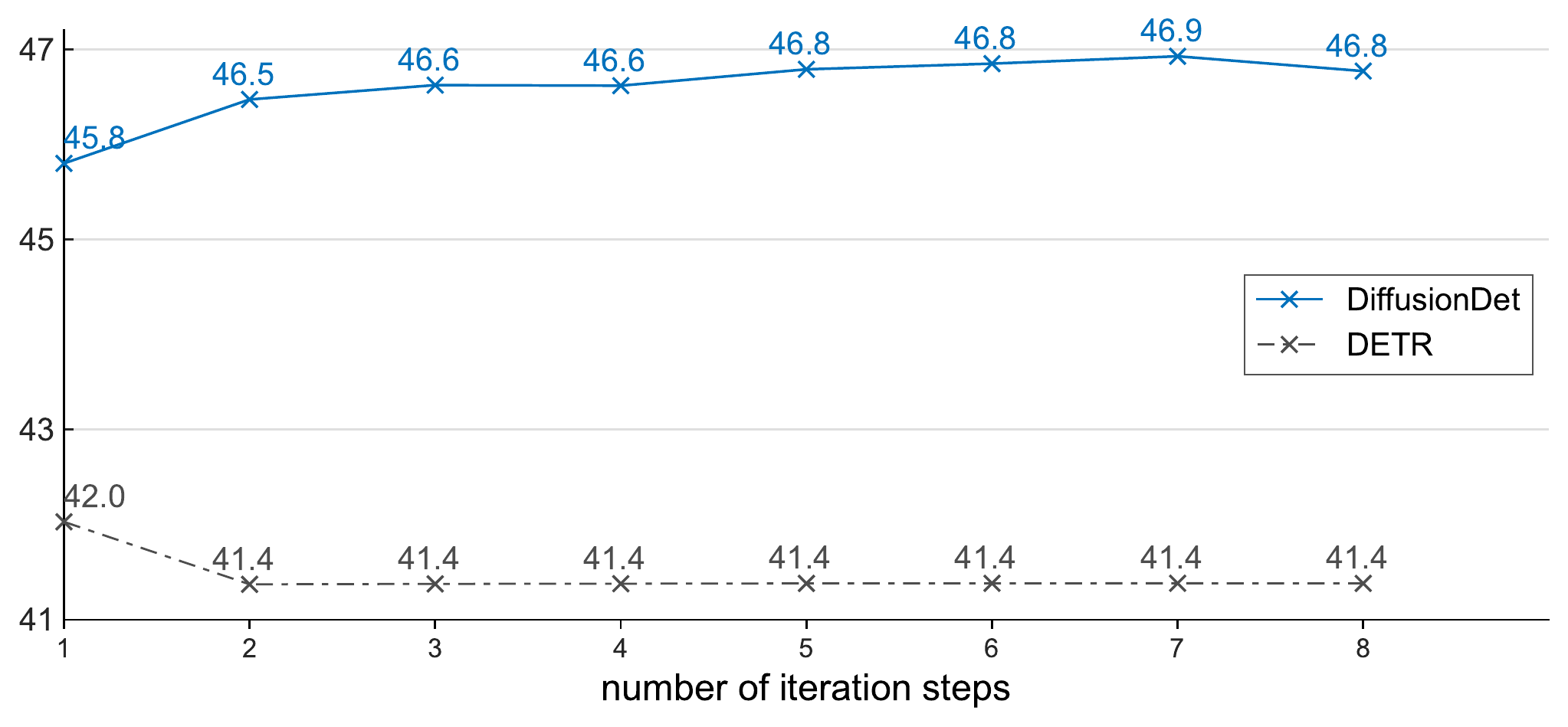}
    \caption{\textbf{DETR~\cite{carion2020end} vs. \ours}}
    \end{subfigure}
    \begin{subfigure}{\linewidth}
    \includegraphics[width=0.98\linewidth]{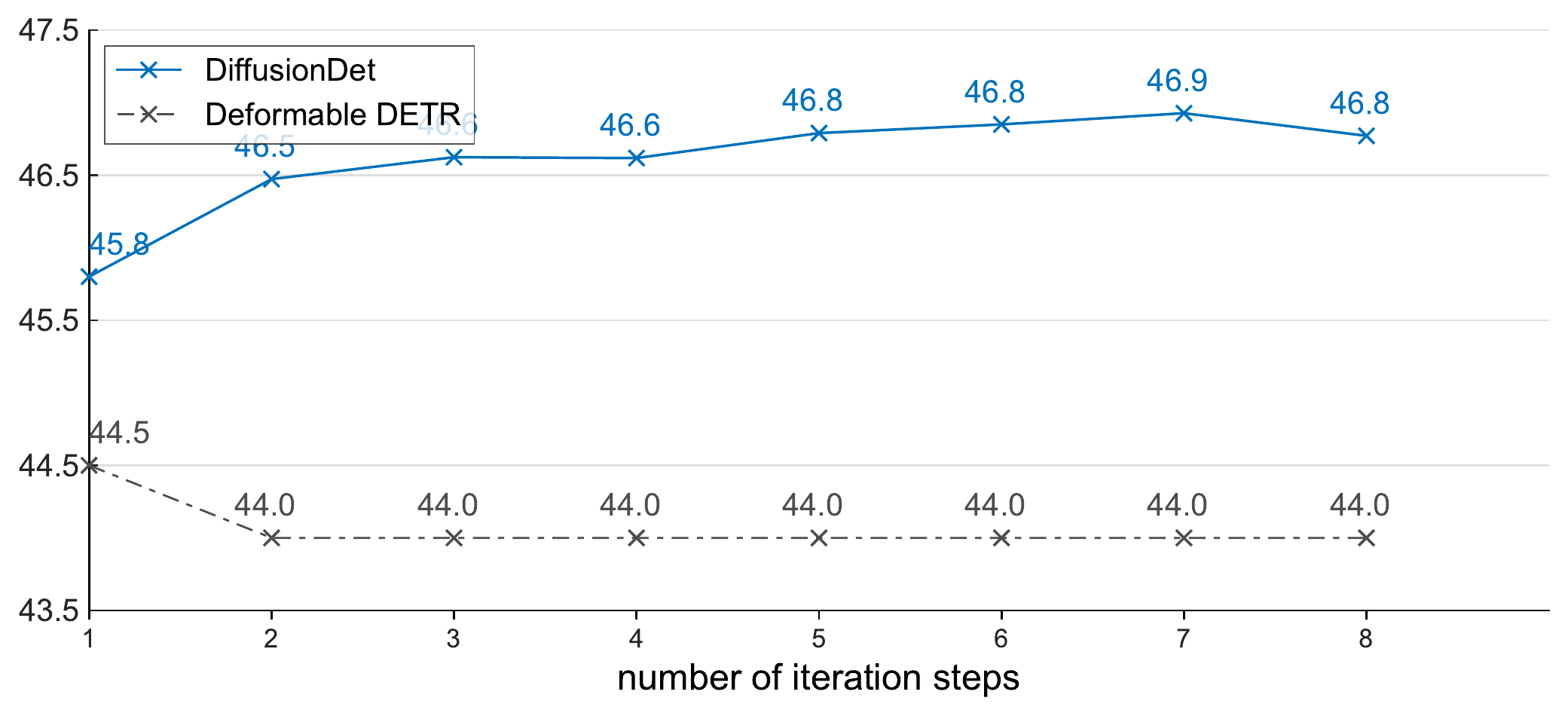}
    \caption{\textbf{Deformable DETR~\cite{zhu2021deformable} vs. \ours}}
    \end{subfigure}
    \begin{subfigure}{\linewidth}
    \includegraphics[width=0.98\linewidth]{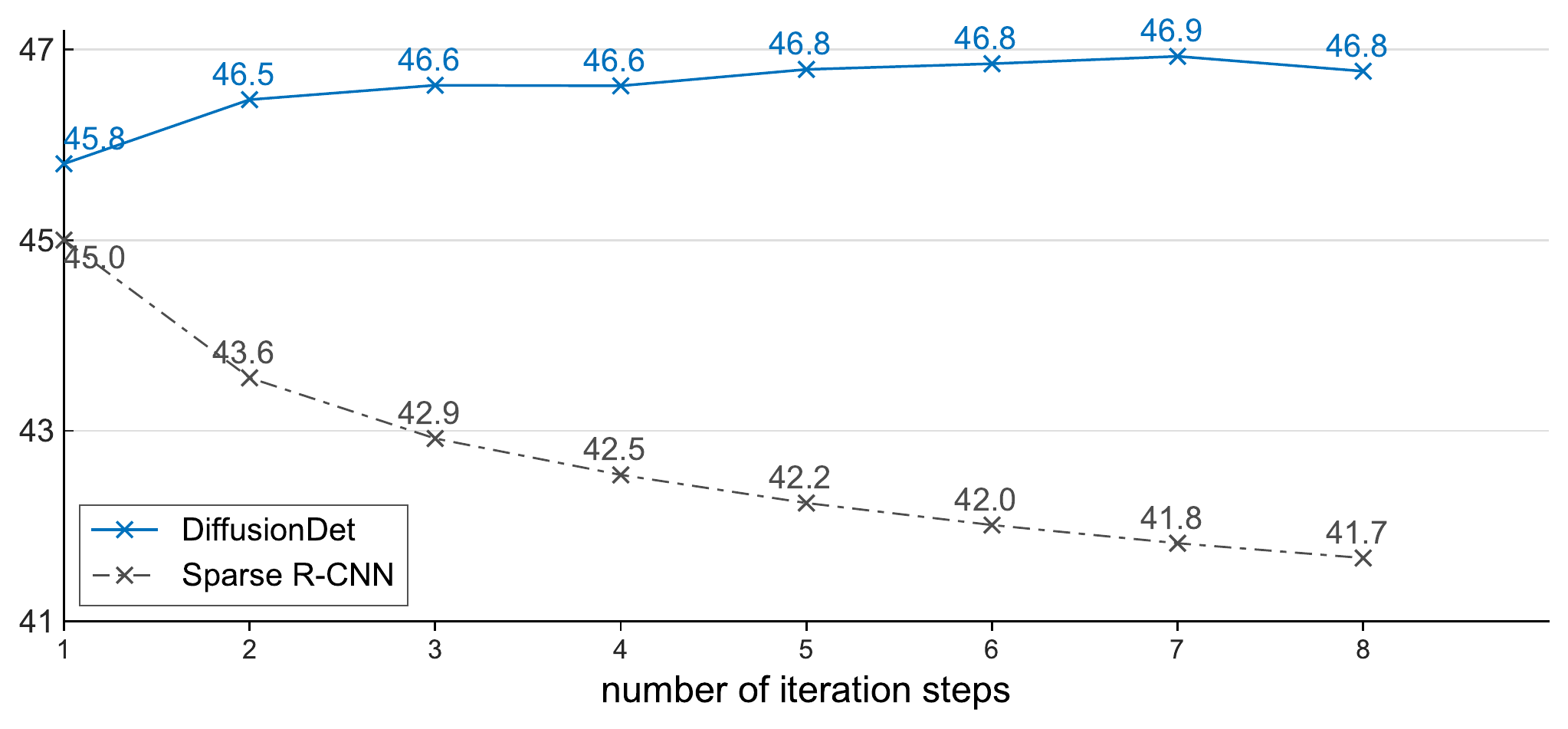}
    \caption{\textbf{Sparse R-CNN~\cite{sun2021sparse} vs. \ours}}
    \end{subfigure}
    \caption{\textbf{Iterative evaluation.} All models are the provided models in their official code repositories.}
    \label{fig:step_refinement}
\end{figure}

We find that all models have performance drop when evaluated with more than one step without ensemble strategy. However, the performance drop of \ours and DETR is negligible. 
Adopting an ensemble would mitigate the performance degradation except for DETR. Nevertheless, previous query-based based methods still have performance drops with more steps. In contrast, \ours turns performance down to up. Specifically, \ours has performance gains with more refinement steps. For example, \ours with five steps has 0.99 AP higher than with a single step. Therefore, we use the ensemble strategy as default. To better compare \ours with DETR, Deformable DETR, and Sparse R-CNN, we also draw the comparison curves of \Cref{tab:ensemble} in \Cref{fig:step_refinement}.

\section{Experimental Settings}\label{sec:app_config}
In this section, we give the detailed experimental settings in \Cref{sec:main_pro} and 
\Cref{sec:benchmark}.

\subsection{DETR with 300 Queries}
Since the official GitHub repository\footnotemark ~of only provides DETR~\cite{carion2020end} with 100 object queries, we reproduce it with 300 object queries using the official code for fair comparison in \Cref{sec:main_pro}. Specifically, we train this model with \texttt{Detectron2} wrapper based on configuration \url{https://github.com/facebookresearch/detr/blob/main/d2/configs/detr_256_6_6_torchvision.yaml}, as summarized in \Cref{tab:impl_detr300}. We note that the configuration only trains the model for about 300 epochs. We only change the \texttt{NUM\_OBJECT\_QUERIES} from 100 to 300 and leave everything else the same as the original one.

\footnotetext{\scriptsize\url{https://github.com/facebookresearch/detr}}

\begin{table}[ht]
\tablestyle{6pt}{1.02}
\begin{tabular}{y{70}|y{140}}
\shline
config & value \\
\shline
\# object queries  & 300 \\
optimizer & AdamW~\cite{loshchilov2018decoupled} \\
base learning rate & 1e-4 \\
weight decay & 1e-4 \\
optimizer momentum & $\beta_1, \beta_2{=}0.9, 0.999$ \\
batch size & 64 \\
learning rate schedule & step lr \\
lr decay steps & (369600,) \\
warmup iter & 10 \\
warmup factor & 1.0 \\
training iters & 554400 \\
clip gradient type &  full model \\
clip gradient value & 0.01 \\
clip gradient norm  & 2.0 \\
data augmentation & \scriptsize{RandomFlip, RandomResizedCrop, RandomCrop}  \\
\shline
\end{tabular}

\vspace{-10pt}
\caption{\textbf{DETR reproduction setting.}}
\label{tab:impl_detr300}
\vspace{-8pt}
\end{table}

\begin{table}[ht]
\tablestyle{6pt}{1.02}
\begin{tabular}{y{70}|y{140}}
\shline
config & value \\
\shline
optimizer & AdamW~\cite{loshchilov2018decoupled} \\
base learning rate & 2.5e-5 \\
weight decay & 1e-4 \\
optimizer momentum & $\beta_1, \beta_2{=}0.9, 0.999$ \\
batch size & 16 \\
learning rate schedule & step lr \\
lr decay steps & (350000, 420000) \\
warmup iter & 1000 \\
warmup factor & 0.01 \\
training iters & 450000 \\
clip gradient type &  full model \\
clip gradient value & 1.0 \\
clip gradient norm  & 2.0 \\
data augmentation & \scriptsize{RandomFlip, RandomResizedCrop, RandomCrop}  \\
\shline
\end{tabular}

\vspace{-10pt}
\caption{\textbf{COCO setting.}}
\label{tab:impl_coco}
\vspace{-8pt}
\end{table}

\begin{table}[ht]
\tablestyle{6pt}{1.02}
\begin{tabular}{y{70}|y{140}}
\shline
config & value \\
\shline
optimizer & AdamW~\cite{loshchilov2018decoupled} \\
base learning rate & 2.5e-5 \\
weight decay & 1e-4 \\
optimizer momentum & $\beta_1, \beta_2{=}0.9, 0.999$ \\
batch size & 16 \\
learning rate schedule & step lr \\
lr decay steps & (210000, 250000) \\
warmup iter & 1000 \\
warmup factor & 0.01 \\
training iters & 270000 \\
clip gradient type &  full model \\
clip gradient value & 1.0 \\
clip gradient norm  & 2.0 \\
data augmentation & \scriptsize{RandomFlip, RandomResizedCrop, RandomCrop}  \\
data sampler & RepeatFactorTrainingSampler \\
repeat thres. & 0.001 \\
\shline
\end{tabular}

\vspace{-10pt}
\caption{\textbf{LVIS setting.}}
\label{tab:impl_lvis}
\vspace{-8pt}
\end{table}

\subsection{Benchmark on COCO and LVIS}
In \Cref{sec:benchmark}, we benchmark \ours on COCO dataset and LVIS dataset. The training configuration is in \Cref{tab:impl_coco} and \Cref{tab:impl_lvis}, respectively.

\section{Training Loss}
We adopt set prediction loss~\cite{carion2020end, sun2021sparse, zhu2021deformable} on the set of $N_{train}$ predictions for \ours. Set prediction loss requires pairwise matching cost between predictions and ground
truth objects, taking into account both the category and box predictions. The matching cost is formulated as :
\begin{equation}
\mathcal{C} = \lambda_{cls}  \cdot  \mathcal{C}_{\mathit{cls}} + \lambda_{L1} \cdot \mathcal{C}_{\mathit{L1}} +
    \lambda_{giou} \cdot \mathcal{C}_{\mathit{giou}},
\end{equation}
where $\mathcal{C}_{\mathit{cls}}$ the focal loss~\cite{lin2017focal} between prediction and ground truth class labels. Besides, our boxes loss contains  $\mathcal{C}_{\mathit{L1}}$ and $\mathcal{C}_{\mathit{giou}}$, which are most commonly-used $\ell_1$ loss and generalized IoU~(GIoU) loss~\cite{rezatofighi2019generalized}. $\lambda_{cls}$, $\lambda_{L1}$ and $\lambda_{giou}$ $\in\mathbb{R}$ are weights of each component to balance to overall multiple losses. Following~\cite{carion2020end, zhu2021deformable, sun2021sparse}, we adopt $\lambda_{cls} = 2.0$, $\lambda_{L1} = 5.0$ and $\lambda_{giou} = 2.0$.

We assign multiple predictions to each ground truth with the optimal transport approach~\cite{ge2021ota, ge2021yolox}.
Specifically, for each ground truth, we select the top-k predictions with the least matching cost as its positive
samples, others as negatives. Then, \ours is optimized with a multi-task loss function:
\begin{equation}\label{total_loss}
\mathcal{L} = \lambda_{cls}  \cdot  \mathcal{L}_{\mathit{cls}} + \lambda_{L1} \cdot \mathcal{L}_{\mathit{L1}} +
    \lambda_{giou} \cdot \mathcal{L}_{\mathit{giou}},
\end{equation}
The component of training loss is the same as the matching cost, except that the loss is only performed on the matched pairs.

% \clearpage
%%%%%%%%% REFERENCES

{\small
\bibliographystyle{ieee_fullname}
\bibliography{egbib}
}

\end{document}